\documentclass{article}

    \usepackage[preprint]{neurips_2024}

\usepackage[utf8]{inputenc} 
\usepackage[T1]{fontenc}    
\usepackage{hyperref}       
\usepackage{url}            
\usepackage{booktabs}       
\usepackage{amsfonts}       
\usepackage{nicefrac}       
\usepackage{microtype}      
\usepackage{xcolor}         

\usepackage{times}
\usepackage{soul}
\usepackage{url}
\usepackage[utf8]{inputenc}
\usepackage[small]{caption}
\usepackage{graphicx}
\usepackage{amsmath}
\usepackage{amsthm}
\usepackage{booktabs}
\usepackage{algorithm}
\usepackage{algorithmic}
\usepackage[switch]{lineno}
\usepackage{multirow}
\usepackage{multicol}
\usepackage{subcaption}
\usepackage{arydshln}
\usepackage[normalem]{ulem}
\usepackage{caption}

\title{In-Context Learning State Vector with Inner and Momentum Optimization}

\author{
    Dongfang Li,
    Zhenyu Liu,
    Xinshuo Hu,
    Zetian Sun,
    Baotian Hu\thanks{Corresponding author.},
    Min Zhang \\
 Harbin Institute of Technology (Shenzhen), Shenzhen, China\\
\texttt{\{crazyofapple, liuzhenyuhit\}@gmail.com} \\
\texttt{\{hubaotian, zhangmin2021\}@hit.edu.cn}
}

\begin{document}

\maketitle

\begin{abstract}
  Large Language Models (LLMs) have exhibited an impressive ability to perform In-Context Learning (ICL) from only a few examples. Recent works have indicated that the functions learned by ICL can be represented through compressed vectors derived from the transformer. However, the working mechanisms and optimization of these vectors are yet to be thoroughly explored. In this paper, we address this gap by presenting a comprehensive analysis of these compressed vectors, drawing parallels to the parameters trained with gradient descent, and introducing the concept of state vector. Inspired by the works on model soup and momentum-based gradient descent, we propose inner and momentum optimization methods that are applied to refine the state vector progressively as test-time adaptation. Moreover, we simulate state vector aggregation in the multiple example setting, where demonstrations comprising numerous examples are usually too lengthy for regular ICL, and further propose a divide-and-conquer aggregation method to address this challenge. We conduct extensive experiments using Llama-2 and GPT-J in both zero-shot setting and few-shot setting. The experimental results show that our optimization method effectively enhances the state vector and achieves the state-of-the-art performance on diverse tasks.
  Code is available at \texttt{https://github.com/HITsz-TMG/ICL-State-Vector}
\end{abstract}

\section{Introduction}

In-Context Learning (ICL) has emerged as a powerful capability in tandem with the scaling of large language models (LLMs)~\citep{gpt3}. By simply conditioning on a few input-label pairs as demonstrations, LLMs yield a significant improvement and deliver remarkable results in various downstream Natural Language Processing (NLP) tasks~\citep{DBLP:journals/tmlr/WeiTBRZBYBZMCHVLDF22,liu2023pre}.
For example, a model prompted with the input \textit{``gaot $\rightarrow$ goat, sakne $\rightarrow$ snake, brid $\rightarrow$''} can produce the output \textit{``bird''}. 
Given these successes, it is worthwhile to inquire about the exact internal working mechanisms of ICL. 
Considering the opaque operation of ICL within the auto-regressive transformer, it is plausible that ICL might function as a general mechanism that leverages both demonstrations and the query to yield the prediction~\citep{dong2022survey}.


Recently, some studies have found that the ICL mapping function exists in the outputs of the attention layers or attention heads~\citep{Liu2023IncontextVM,ICL_Gradient_Descent_as_Meta_Optimizers} when applying causal effects analysis on a different set of models and tasks, such as the task vector~\citep{Hendel2023InContextLC} and the function vector~\citep{Todd2023FunctionVI}.
These works show that the functionalities learned through ICL can be encapsulated in compressed vectors derived from transformers, which then can be used to intervene in the transformer to handle queries without demonstrations. 
This revelation suggests the potential mechanism of ICL that first utilises demonstrations to learn the mapping function from inputs to labels in shallow transformer layers, and then uses the ICL function in deeper transformer layers to make predictions~\citep{Hendel2023InContextLC}.
However, while these compressed vectors encapsulate learned information in a more condensed form and show significant promise in applying ICL, there still exists a considerable gap in understanding the operational mechanisms and optimization strategies of these vectors. This significant gap hinders the further grasping and utilization of ICL.


In this paper, we aim to bridge the existing gap by presenting a comprehensive analysis of compressed vectors. Specifically, we investigate their similarities with parameters trained via gradient descent and introduce the formulation of state vector that encapsulates the processing state of ICL stored in the attention activations. Building on the concept of state vector, and drawing insights from the model soup~\citep{Wortsman2022ModelSA} and momentum-based gradient optimization algorithms~\citep{momentum_ref,sgd}, we propose inner optimization and momentum optimization strategies which are progressively applied to enhance the state vector.
Moreover, we further exploit the demonstration compression capabilities of the state vector to address the practical challenges encountered when applying ICL in settings with multiple examples, where demonstrations are typically too lengthy for standard ICL, such as in the 100-shot setting which is prevalent in practice. Specifically, we introduce a divide-and-conquer aggregation method that effectively aggregates the ICL functions of these extensive examples. This approach enables us to scale up for processing extended examples by compressing them into a single state vector.
We conduct extensive experiments using Llama-2~\citep{Touvron2023Llama2O} and GPT-J~\citep{gpt-j} in both zero-shot and few-shot settings. The experimental results show that our method effectively enhances the state vector and achieves state-of-the-art performance on diverse tasks. 
This not only manifests the effectiveness of our approach but also paves the way for a more comprehensive understanding of ICL.


Our contributions are summarized as follows:
\begin{itemize}
\item We delve into the working mechanism of compressed vectors in ICL and highlight their similarities with parameters trained via gradient descent. Building on this observation, we propose the formulation of the state vector.
\item 
We propose inner and momentum optimization to progressively refine the state vector as an efficient test-time adaptation. Additionally, we introduce a divide-and-conquer aggregation to effectively scale up to large numbers of examples.
\item We show the practicality of our proposed methods across a wide range of tasks through extensive experiments. Our results also offer insights for future research aiming to fully understand the functionalities of ICL.
\end{itemize}

\section{Related Work}


\paragraph{Mechanistic Interpretability.} Recent works have focused on the working mechanisms of ICL~\citep{DBLP:conf/nips/ChanSLWSRMH22,DBLP:conf/iclr/XieRL022,DBLP:conf/emnlp/WangLDCZMZS23}.~\cite{ICL_Induction_Heads} argue that induction heads may be the mechanistic source of general ICL in transformers.~\cite{Akyrek2022WhatLA} show that transformer-based in-context learners can implicitly implement standard optimization algorithms on linear models. A mainstream assumption posits that ICL has a similarity with the gradient descent.~\cite{Oswald2022TransformersLI} demonstrate how a linear attention-only transformer model can perform a gradient descent-like procedure implicitly.~\cite{ICL_Gradient_Descent_as_Meta_Optimizers} compare standard gradient descent based fine-tuning and ICL, and figure out that the transformer attention of ICL exhibits a dual form of gradient descent-based optimization. Moreover, some works revisit and modify this theory on the layer causality dependence~\citep{Natan2023IncontextLA} or training batch size~\citep{Shen2023DoPT}. 
In contrast, we focus on the application of the dual form of gradient descent and ICL and present optimization methods with inspiration from the dual form.
\paragraph{Task Representation.} 
Numerous studies have extensively explored the concept of compressing various tasks into task representations as a means of effectively manipulating tasks within ICL ability.
Notably,~\cite{Shao2023CompositionalTR} and~\cite{Mu2023LearningTC} have successfully yielded compositional task representations by training a composition model. In a slightly different vein, some researchers have delved into the art of devising methodologies to compose minor parameter adjustments acquired through task fine-tuning~\citep{Ilharco2022EditingMW,Panigrahi2023TaskSpecificSL,Yu2023LanguageMA,hu2023separate,DBLP:journals/corr/abs-2305-16130}. 
An alternative line of research finds that the task representation could be extracted in ICL~\citep{Liu2023IncontextVM,Hendel2023InContextLC,Todd2023FunctionVI,DBLP:journals/corr/abs-2305-13016}. 
Different from these approaches, our work avoids the need for additional training and focuses more on analysing why these compressed vectors work and how to improve their performance.

\section{Formalization}
\label{sec:TF}

In this section, we first provide a detailed examination of attention activation which is found to contain the compressed ICL function by previous works~\citep{Hendel2023InContextLC,Todd2023FunctionVI}. Then, we highlight its inherent similarities with parameters trained through gradient descent. Finally, we introduce the concept of the state vector drawing inspiration from these observations.

A classic template of ICL has the following necessary components:
(1) $N$ examples that are used to form the demonstrations and each example contains an input query $\mathcal{X}$ and its corresponding label $\mathcal{Y}$.
(2) Separate tokens $\mathcal{S}$ that separate the input query and the label for each example (e.g., $\rightarrow$). 
(3) A query $\mathcal{X}_q$ for prediction. 
With the above components, the contextual model input of ICL could be written as follows:
\begin{equation*}
    \mathcal{X}_1, \mathcal{S}, \mathcal{Y}_1, \mathcal{X}_2, \mathcal{S}, \mathcal{Y}_2, \cdots, \mathcal{X}_N, \mathcal{S},  \mathcal{Y}_N, \mathcal{X}_q, \mathcal{S}.
\end{equation*}
Here we analyse the attention activation of the last separate token.
In the $l$-th transformer layer, the output activation $\mathbf{a}^l$ of the attention heads of the last separate token is:
\begin{equation}
    \mathbf{a}^l = W_{V}[X^{\prime}; X] \operatorname{softmax} \left( \frac{\left( W_{K} [X^{\prime}; X] \right)^T  \mathbf{q}}{\sqrt{d}} \right),
\end{equation}
where $X^{\prime}$ denotes the hidden state of demonstrations, $X$ denotes the hidden state of the query and the last separate token (called zero-shot input), $q$ denotes the attention query vector of the last separate token, $[X^{\prime}; X]$ denotes the matrix concatenation, $\sqrt{d}$ is the scaling factor, $W_K$ and $W_V$ are parameter weight matrix.

Consistent with previous works~\citep{ICL_Gradient_Descent_as_Meta_Optimizers,Natan2023IncontextLA}, we omit the softmax operation and the scaling factor to approximate standard attention as relaxed linear attention for qualitative analysis.
Consequently, the activation can be simplified as follows:
\begin{equation}
    \begin{aligned}
    \mathbf{a}^l & \approx W_{V} [X^{\prime}; X] \left( W_{K} [X^{\prime}; X] \right)^T \mathbf{q} \\
    & = \left( W_{V} X \left( W_{K} X \right)^T + W_{V} X^{\prime} \left( W_{K} X^{\prime} \right)^T \right) \mathbf{q} \\
    & = \left( W_{\text{ZSL}} + \sum_i \left( (W_{V} \textbf{x}^{\prime}_i) \otimes \left( W_{K} \textbf{x}^{\prime}_i \right) \right) \right) \mathbf{q}.
    \end{aligned}
    \label{equ:task vector zsl}
\end{equation}
We define $W_{\text{ZSL}}=W_{V} X \left( W_{K} X \right)^T$ as the initialized parameters since it is the attention result in the Zero-Shot Learning (ZSL) setting.

To draw a meaningful comparison between attention activation and parameters trained through gradient descent, we now shift our focus towards analyzing a simple linear transformation represented by $\mathbf{y}_i=W\mathbf{x}_i$. Given a loss function $\mathcal{L}$ and the learning rate $\eta$, the gradient of linear weight is:
\begin{equation}
   \nabla_W \mathcal{L}(\mathbf{y}_i) = \frac{\partial \mathcal{L}(\mathbf{y}_i)}{\partial{\mathbf{y}_i}}  \frac{\partial \mathbf{y}_i}{\partial W} = \nabla_{\mathbf{y}_i} \mathcal{L}(\mathbf{y}_i) \mathbf{x}_i^T.
\end{equation}
Denoting the back-propagated errors as $\mathbf{e}_i = -\eta \nabla_{\mathbf{y}_i}\mathcal{L}$, we can get the full batch gradient with training examples:
\begin{equation}
    \Delta W_{GD} = \sum_i \mathbf{e}_i \otimes \mathbf{x}^{\prime}_i,
\end{equation}
where $\mathbf{x}^{\prime}_i$ is the input training examples. Hence, in the previous Eqn.~\ref{equ:task vector zsl}, if we substitute $W_{K} \textbf{x}^{\prime}_i$ as training examples, and take $W_{V} \textbf{x}^{\prime}_i \approx \mathbf{e}_i$ corresponding to some meta gradients~\citep{ICL_Gradient_Descent_as_Meta_Optimizers,Natan2023IncontextLA}. 
The activation can be written as:
\begin{equation}
    \mathbf{a}^l  = \left( W_{\text{ZSL}} + \sum_i \mathbf{e}_i \otimes W_{K} \textbf{x}^{\prime}_i \right) \mathbf{q} 
    = \left( W_{\text{ZSL}} + \Delta W_{GD} \right) \mathbf{q}.
    \label{equ:tv as gd}
\end{equation}
Hence, it can be inferred that the output activation $\mathbf{a}^l$ can be regarded as parameters trained via gradient descent which utilizes the demonstrations as training instances. 

With the above dual form between activation and trained parameters, and in light of observations that transformers tend to learn the ICL function primarily in their first $L$ layers~\citep{DBLP:conf/emnlp/WangLDCZMZS23}, we have the following hypothesis:
During the process of ICL, the first $L$ layers progressively update the flow of information using each example in the demonstration through forward computation. The processing state of ICL is then stored within the activation of the attention head. The subsequent layers access and utilize the processing state to reinstate the ICL function, which is used implicitly for predicting the queries.
Therefore we concatenate the activation in the initial $L$ layers and introduce the notation of the state vector:
\begin{equation}
    \mathcal{V}^L_N = \mathop{\Big{\|}}\limits_{l=1}^{L} \mathbf{a}^l,
\end{equation}
where $L$ is the number of layers and $N$ is the number of examples in the demonstration. $\|$ denotes the concatenation operation.
Note that we have a completely different construction strategy and usage compared to the function vector~\citep{Todd2023FunctionVI}. Although the task vector~\citep{Hendel2023InContextLC} may be functionally equivalent in the forward process, the proposed state vector differs significantly in its integration into the model, making it easier and more effective to analyse and interpret. 

\section{Method}
\label{sec:Method}
\begin{figure*}[htb]
    \centering
    \includegraphics[width=\textwidth]{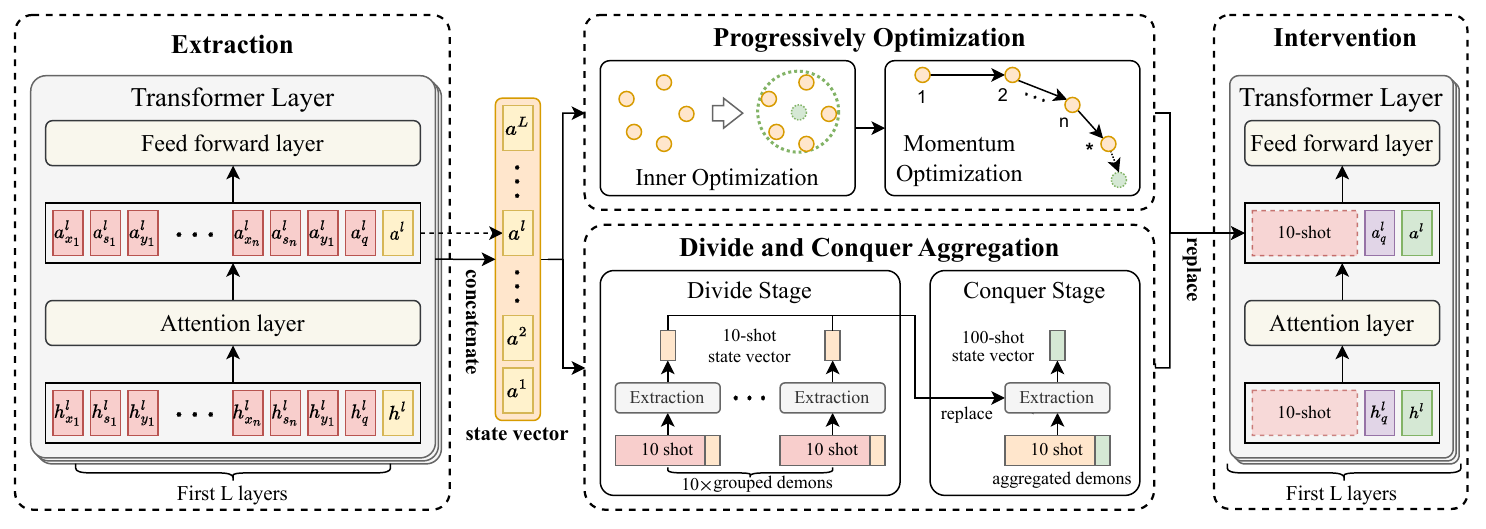}
    \caption{
    The overall framework of the proposed state vector. 
    The state vectors are extracted from the output activations of attention heads.
    These state vectors are progressively optimized by \textit{inner optimization} and \textit{momentum optimization}, or be aggregated through a \textit{divide-and-conquer (D\&C) aggregation}.
    Finally, the processed state vector is utilized to intervene the inference forward pass.
    }
    \label{fig:framework}
\end{figure*}

\subsection{Overview}
\label{sec:Basic Inference}

As illustrated in Figure~\ref{fig:framework}, our approach initially extracts the state vector from the attention head that corresponds to the final separate token in the first $L$ layers using a demonstration and a dummy query.
Then, with the view of treating the state vector as trained parameters, coupled  with drawing inspiration from the model soup and the momentum-based gradient optimization algorithm, we introduce two methods that progressively optimize the state vector as test-time adaptation~\citep{DBLP:journals/corr/abs-2303-15361}: (1) inner optimization (\S\ref{sec:Average Optimize}) and (2) momentum optimization (\S\ref{sec:Momentum Optimize}). 
Moreover, we propose a divide-and-conquer (D\&C) state vector aggregation method for efficiently compressing the ICL function in the multiple example setting (\S\ref{sec:State Vector Aggregation}).

After the state vector optimization or aggregation, we utilize the processed state vector to intervene the model during the forward inference pass. In particular, we first input a test query in the zero-shot setting or with the demonstration in the few-shot setting. During the forward pass in the first $L$ layers, we replace the attention activation of the last separate token with the corresponding activation in the state vector. In other words, the state vector is leveraged to intervene in the output of the first $L$ transformer layers, blocking the attention of the last separate token to the previous context. 
With state vector intervention, the transformer learns the ICL function from the processing state stored in the state vector, and continues to make the prediction on the test query.

\subsection{Inner Optimization}
\label{sec:Average Optimize}
Inspired by the works on the model soup~\citep{Wortsman2022ModelSA,DBLP:conf/eacl/ChronopoulouPFD23} which show that weight-space averaging not only yields performance improvement but also often enhances robustness, we thus ask the following research question (\textbf{RQ1}): \textit{Is it possible to optimize our state vector using the model soup approach?}
To explore this question, we propose an inner optimization method to improve the effectiveness and robustness of state vector. Specifically, we not only extract the state vector in each separate token of the dummy query but also extract the state vector from each example. 
Formally, with a forward pass in an $N$ shot ICL setting, we extract the $N$ state vector $\mathcal{V}^L_{i}$ ($1 \leq i \leq N$) from last $N$ separate token. Subsequently, we apply a uniform averaging process to these state vectors as follows:
\begin{equation}
    \mathcal{\overline{V}}^L_N = \frac{1}{N} \sum^N_{i=1} \mathcal{V}^L_i,
\end{equation}
where $\mathcal{\overline{V}}^L_N$ is the inner optimized state vector, which can be directly utilized for inference intervention or serves as the initial state vector for later momentum optimization.
 
\subsection{Momentum Optimization}
\label{sec:Momentum Optimize}
Since we view the state vector as parameters trained gradually through demonstration examples, the difference between two state vectors with adjacent corresponding separate tokens can also be regarded as the influence of the middle example, akin to the gradient. Motivated by this understanding, coupled with extensive studies of the gradient optimization algorithm~\citep{sgd,adagrad,adamw}, we direct our focus toward a simple momentum-based gradient optimization algorithm, seeking to answer the following research question (\textbf{RQ2}): \textit{Can our state vector be optimized using momentum-based optimization algorithm?} 
To answer this question, we propose a momentum optimization.
Formally, we first extract the influence of each example by subtracting two adjacent state vectors:
\begin{equation}
    E^L_i = \mathcal{V}^L_{i} - \mathcal{V}^L_{i-1},
\end{equation}
where $E^L_i$ is the influence of $i$-th ($1 < i \leq N$) example in the early $L$ layer. Then, we apply the momentum gradient optimization algorithm to obtain optimized influence $\widetilde{E}^L_i$, and add it to the last state vector:
\begin{equation}
\label{grad_eqn}
    \mathcal{\widehat{V}}^L_N = \mathcal{\overline{V}}^L_N + \widetilde{E}^L  
    = \mathcal{\overline{V}}^L_N + \texttt{opt}([E^L_i]_{i=1}^{N}),
\end{equation}
where $\mathcal{\widehat{V}}^L_N$ is the momentum optimized state vector and $\mathcal{\overline{V}}^L_N$ is the inner optimized state vector. $\texttt{opt}(\cdot)$ denotes the momentum gradient optimization algorithm. We also explore various other gradient optimization algorithms in \S\ref{sec:GOA}.

\subsection{Divide-and-Conquer Aggregation}
\label{sec:State Vector Aggregation}

In addition to optimizing the state vector to more effectively represent the ICL function from a small number of examples, we also explore its capacity to encapsulate multiple examples within a single vector. However, regular ICL can not be directly used on multiple examples due to the context length limitation of current LLMs. This leads us to investigate the following question (\textbf{RQ3}): 
\textit{Can we use the state vector to represent multiple examples that are unmanageable for regular ICL?} 
To address this question, we propose a divide-and-conquer method for state vector aggregation. As depicted in Figure~\ref{fig:framework}, our approach involves distinct  aggregation processes (i.e. the divide stage and the conquer stage). 
In the divide stage, examples are randomly divided into groups, termed grouped demonstrations. Within each group, a random example is selected to serve as a dummy query, which allows us to extract a group-specific state vector. 
In the conquer stage, these dummy queries are paired with their corresponding labels to form input-label pairs. From these input-label pairs, we form an aggregated demonstration, add an additional dummy query, and subsequently extract the aggregated state vector.
It is worth noting that during the forward pass of aggregated state vector extraction, we utilise the group-specific state vector to intervene the attention activation of the separate tokens of their corresponding examples.
The divide and conquer approach allows us to aggregate the ICL function of each grouped demonstration into its respective group-specific state vector, and subsequently aggregate the ICL function of each group-specific state vector into a single, comprehensive aggregated state vector. This aggregated vector is then utilized for interventions during inference, similarly to the optimized state vector discussed in \S4.2 and \S4.3. Moreover, in the few-shot setting, the aggregated demonstrations are treated as inference demonstrations.
The divide-and-conquer approach effectively circumvents the context-length constraints inherent in LLMs, thereby enabling a more effective and efficient aggregation of information across multiple examples.


\section{Experiment}

\subsection{Setup}

\label{sec:experimental setup}

We conduct the evaluation across 12 datasets that encompass different domains.
\begin{itemize}
\item  \textbf{Linguistics} includes Antonym~\citep{Nguyen2017DistinguishingAA}, Capitalize, Present-Past, and Singular-Plural~\citep{Todd2023FunctionVI}, focusing on transformations in the form or meaning of words.
\item \textbf{Translation} is represented by the English-French~\citep{Conneau2017WordTW} dataset, which involves translating English words into their French counterparts.
\item \textbf{Knowledge} comprises Country-Capital~\citep{Todd2023FunctionVI}, AG News~\citep{DBLP:conf/nips/ZhangZL15}, Person-Sport, Person-Instrument, Person-Occupation, Product-Company, and Landmark-Country~\citep{Hernandez2023LinearityOR}, which are centred around question-to-answer mappings for commonsense knowledge queries.
\end{itemize}

We employ \emph{Llama-2-7B} and \emph{GPT-J-6B} as our LLMs, chosen for their moderate model sizes, open-source and capability for ICL. We also provide the results with larger models (i.e., Llama-2-13B) in the Appendix~\ref{sec: larger model}. We use Llama-2-7B as the default model unless otherwise specified.
Our method is orthogonal to the choice of transformer-based decoder-only autoregressive LLMs. 

For simplicity evaluation, we restrict to single-token output and use first output token accuracy as the evaluation metric as in previous work~\citep{Hendel2023InContextLC,Todd2023FunctionVI}.

\subsection{Baseline}
\label{sec:Baseline}
In the paper, we compare with the following methods:
\begin{itemize} 
    \item \textbf{Regular} is the baseline for the zero-shot setting that uses only the given query as input, while \textbf{ICL baseline}~\citep{DBLP:journals/tmlr/WeiTBRZBYBZMCHVLDF22} makes predictions on the label by taking both the demonstrations and the given query.
    \item \textbf{Function vector}~\citep{Todd2023FunctionVI} is extracted from attention activation using the causal mediation method and is then added to the hidden state of certain transformer layers during inference. 
    \item \textbf{Task vector}~\citep{Hendel2023InContextLC} is extracted from the hidden state of the separate token and is leveraged for blocking the layer when inference. 
\end{itemize}

\begin{table*}[ht!]
\centering
\scriptsize
{\renewcommand{\arraystretch}{1.1}%
\setlength{\tabcolsep}{4pt}
	\begin{tabular}{c|l|l|ccc ccc|c}
		\hline
        \textbf{Model} &\multicolumn{2}{c|}{\textbf{Method}} &\textbf{Anym}  &\textbf{Eng-Fr} &\textbf{Pers-Inst}  &\textbf{Pers-Occ}  &\textbf{Prod-Comp}    &\textbf{Land-Cout}  &\textbf{Average}\\
        \hline       
        \multirow{10}{*}{Llama-2}
            &\multirow{5}{*}{Zero-shot}       
                &Regular                    &1.0\tiny{$\pm$ 0.2}	            &0.1\tiny{$\pm$0.1}	            &0.0\tiny{$\pm$0.0}	            &0.0\tiny{$\pm$0.0}	            &0.4\tiny{$\pm$0.2}	            &0.0\tiny{$\pm$0.0}             &0.3     \\
            &   &Function vector            &45.1\tiny{$\pm$2.0}	            &21.6\tiny{$\pm$2.0}	        &11.3\tiny{$\pm$10.7}           &0.1\tiny{$\pm$0.1}             &25.6\tiny{$\pm$4.3}	        &32.9\tiny{$\pm$21.6}           &22.8    \\
            &   &Task vector                &56.2\tiny{$\pm$2.8}                &63.2\tiny{$\pm$3.6}            &61.8\tiny{$\pm$8.4}            &27.9\tiny{$\pm$15.2}           &55.5\tiny{$\pm$20.1}           &57.8\tiny{$\pm$26.3}           &53.7    \\
            &   &State vector (inn.)        &\uline{61.0}\tiny{$\pm$1.0}       &66.5\tiny{$\pm$2.2}            &67.4\tiny{$\pm$2.6}            &42.7\tiny{$\pm$4.2}            &64.5\tiny{$\pm$10.6}           &\uline{81.0}\tiny{$\pm$1.7}   &63.9    \\
            &   &State vector (mom.)        &60.4\tiny{$\pm$0.7}                &\uline{67.5}\tiny{$\pm$1.8}   &\uline{68.7}\tiny{$\pm$1.6}   &\uline{45.6}\tiny{$\pm$5.9}   &\uline{71.3}\tiny{$\pm$3.6}   &77.7\tiny{$\pm$1.8}            &\uline{65.2}    \\   
            \cline{2-10}
            &\multirow{5}{*}{Few-shot}     
                &ICL baseline               &64.8\tiny{$\pm$4.8}                &74.3\tiny{$\pm$0.8}            &71.7\tiny{$\pm$3.7}            &56.1\tiny{$\pm$2.7}            &80.8\tiny{$\pm$0.8}            &87.0\tiny{$\pm$0.3}            &72.5    \\
            &   &Function vector            &54.5\tiny{$\pm$0.9}	            &65.2\tiny{$\pm$1.4}	        &60.8\tiny{$\pm$5.6}            &54.2\tiny{$\pm$2.2}	&76.0\tiny{$\pm$1.3}	        &84.2\tiny{$\pm$2.9}            &65.8    \\
            &   &Task vector                &65.7\tiny{$\pm$1.8}                &73.8\tiny{$\pm$0.9}            &66.6\tiny{$\pm$5.2}            &56.4\tiny{$\pm$2.3}            &81.9\tiny{$\pm$1.8}            &86.7\tiny{$\pm$0.9}            &71.8    \\
            &   &State vector (inn.)        &\textbf{66.2}\tiny{$\pm$1.6}       &\textbf{74.6}\tiny{$\pm$0.9}   &70.1\tiny{$\pm$4.3}            &57.0\tiny{$\pm$2.2}            &\textbf{82.8}\tiny{$\pm$1.6}   &87.5\tiny{$\pm$0.9}            &73.0    \\
            &   &State vector (mom.)        &65.8\tiny{$\pm$3.7}                &74.3\tiny{$\pm$1.1}            &\textbf{74.9}\tiny{$\pm$2.9}   &\textbf{58.2}\tiny{$\pm$0.4}   &82.0\tiny{$\pm$1.0}            &\textbf{87.6}\tiny{$\pm$0.3}   &\textbf{73.8}    \\
            \noalign{\hrule height 1pt}
        \multirow{10}{*}{GPT-J} 
            &\multirow{5}{*}{Zero-shot}    
                & Regular                   &8.1\tiny{$\pm$0.6}	                &7.2\tiny{$\pm$0.6}	            &0.0\tiny{$\pm$0.0}	            &0.0\tiny{$\pm$0.0}	            &1.9\tiny{$\pm$0.5}	            &0.9\tiny{$\pm$0.2}             &3.0     \\
            &   & Function vector           &33.1\tiny{$\pm$1.8}	            &29.1\tiny{$\pm$8.5}            &4.1\tiny{$\pm$5.8}	            &11.1\tiny{$\pm$2.3}            &\uline{46.3}\tiny{$\pm$5.7}	&22.5\tiny{$\pm$10.2}           &24.4    \\
            &   &Task vector                &23.6\tiny{$\pm$3.8}                &32.2\tiny{$\pm$5.1}            &44.4\tiny{$\pm$5.0}            &28.3\tiny{$\pm$18.6}           &43.8\tiny{$\pm$5.7}            &41.3\tiny{$\pm$12.3}           &35.6    \\   
            &   &State vector (inn.)        &\uline{33.4}\tiny{$\pm$1.9}       &31.7\tiny{$\pm$3.8}            &49.3\tiny{$\pm$2.0}            &30.0\tiny{$\pm$6.2}            &42.8\tiny{$\pm$4.3}            &\uline{61.9}\tiny{$\pm$1.6}   &41.5    \\
            &   &State vector (mom.)        &31.1\tiny{$\pm$1.0}                &\uline{35.1}\tiny{$\pm$2.4}   &\uline{50.3}\tiny{$\pm$3.0}   &\uline{42.4}\tiny{$\pm$1.5}   &44.2\tiny{$\pm$1.5}            &60.3\tiny{$\pm$0.9}            &\uline{43.9}    \\
            \cline{2-10}
            &\multirow{5}{*}{Few-shot}     
                &ICL baseline               &59.2\tiny{$\pm$1.4}                &69.9\tiny{$\pm$2.0}            &44.7\tiny{$\pm$6.7}            &29.3\tiny{$\pm$1.0}            &62.5\tiny{$\pm$1.0}            &\textbf{69.3}\tiny{$\pm$0.5}   &55.8    \\ 
            &   &Function vector            &56.4\tiny{$\pm$1.9}	            &65.8\tiny{$\pm$1.9}	        &49.1\tiny{$\pm$2.2}	        &30.3\tiny{$\pm$1.9}	        &58.5\tiny{$\pm$3.3}	        &69.2\tiny{$\pm$0.6}            &54.9    \\    
            &   &Task vector                &58.5\tiny{$\pm$1.6}                &70.6\tiny{$\pm$1.2}            &42.3\tiny{$\pm$6.4}            &27.8\tiny{$\pm$3.3}            &66.0\tiny{$\pm$2.6}            &63.1\tiny{$\pm$5.3}            &54.7    \\
            &   &State vector (inn.)        &58.7\tiny{$\pm$2.2}                &\textbf{70.9}\tiny{$\pm$1.3}   &46.5\tiny{$\pm$4.9}            &29.4\tiny{$\pm$1.7}            &\textbf{66.3}\tiny{$\pm$2.1}   &66.4\tiny{$\pm$2.8}            &56.4    \\
            &   &State vector (mom.)        &\textbf{59.6}\tiny{$\pm$1.4}       &70.1\tiny{$\pm$2.2}            &\textbf{51.9}\tiny{$\pm$2.4}   &\textbf{30.4}\tiny{$\pm$1.1}   &63.8\tiny{$\pm$0.8}            &68.6\tiny{$\pm$0.3}            &\textbf{57.4}    \\
        \hline
	\end{tabular}
 }
    \caption{Performance of state vector optimization. The best results in the zero shot setting are in \uline{underline} and the best results in the few shot setting are in \textbf{bold}. The result of basic state vector is mathematically equivalent to task vector. Note that we only present the results across six tasks here and leave the rest in the Appendix. We also report standard deviation and the results are passed with significance test ($p < .05$).}
    \label{tab:main result}
\end{table*}

\begin{figure*}[hpt]
\centering
\scriptsize
\begin{subfigure}[b]{0.24\linewidth}
    \includegraphics[width=\linewidth]{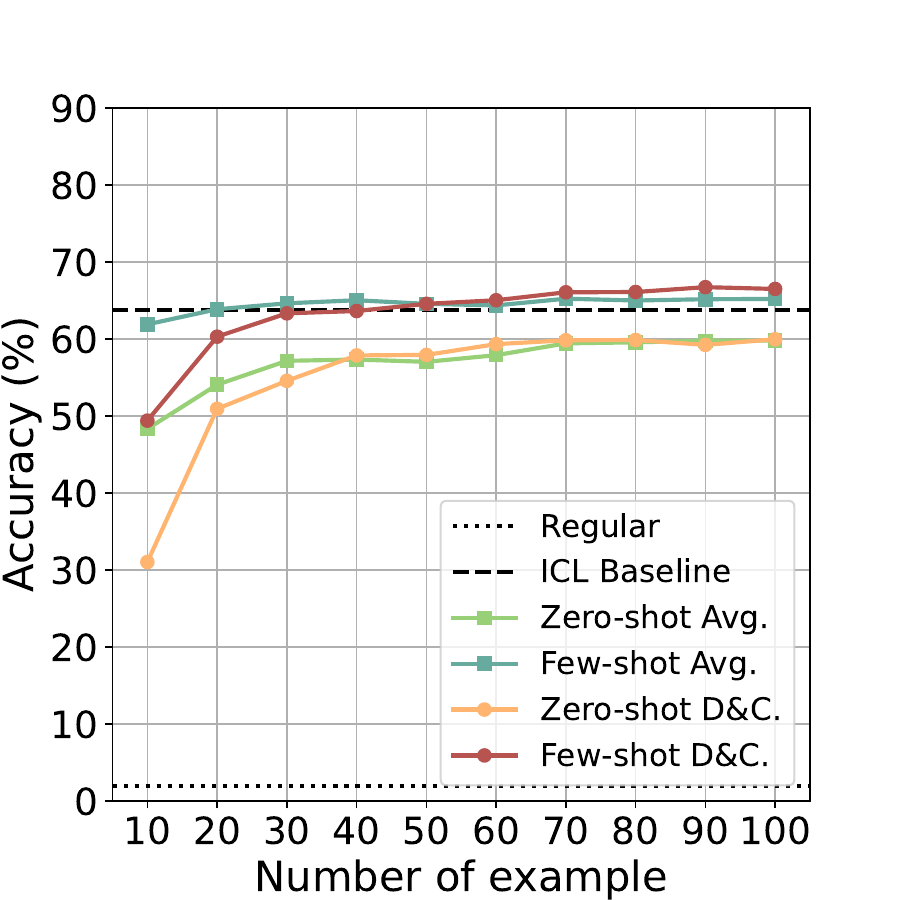}
    \captionsetup{font={scriptsize}}
    \caption{Llama-2 Antonym}
  \end{subfigure}
  \begin{subfigure}[b]{0.24\linewidth}
    \includegraphics[width=\linewidth]{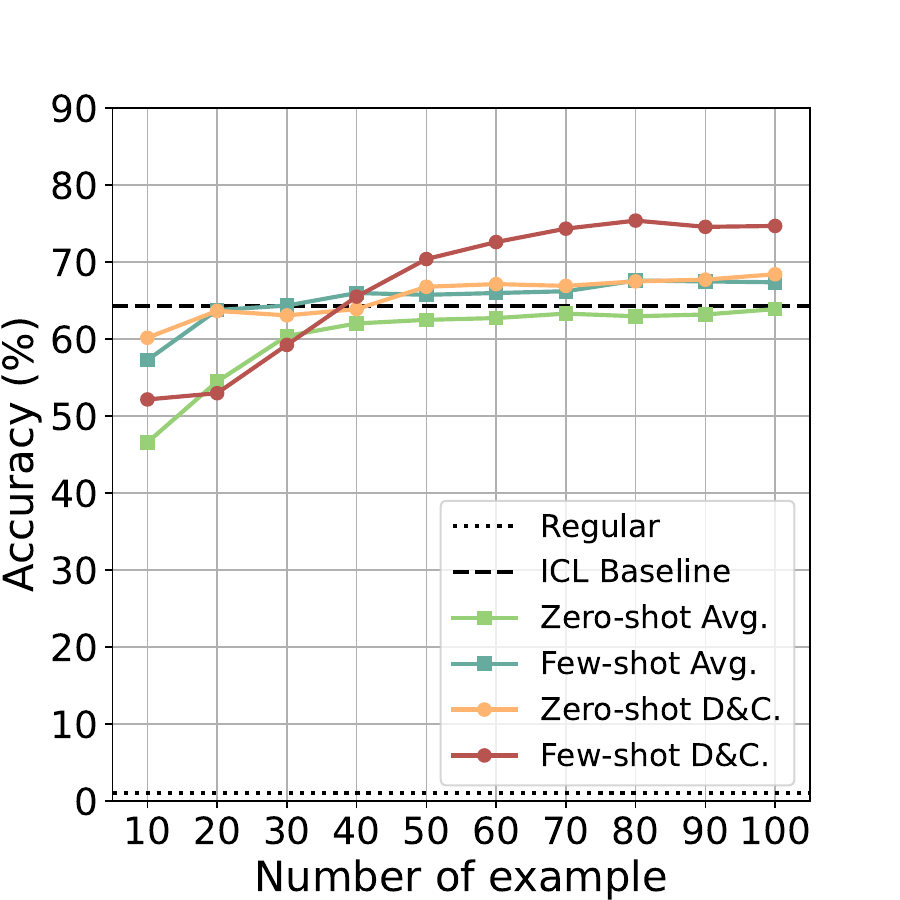}
    \captionsetup{font={scriptsize}}
    \caption{Llama-2 Person-Instrument}
  \end{subfigure}
  \begin{subfigure}[b]{0.24\linewidth}
    \includegraphics[width=\linewidth]{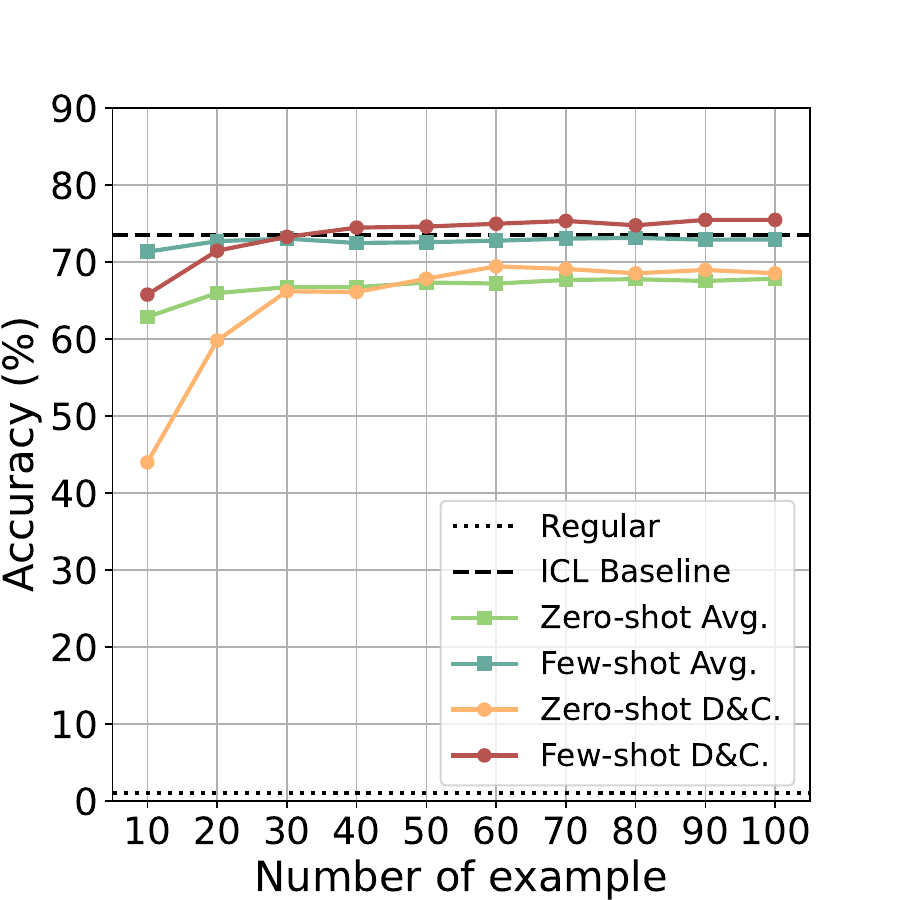}
    \captionsetup{font={scriptsize}}
    \caption{Llama-2 English-French}
  \end{subfigure}
  \begin{subfigure}[b]{0.24\linewidth}
    \captionsetup{font={scriptsize}}
    \includegraphics[width=\linewidth]{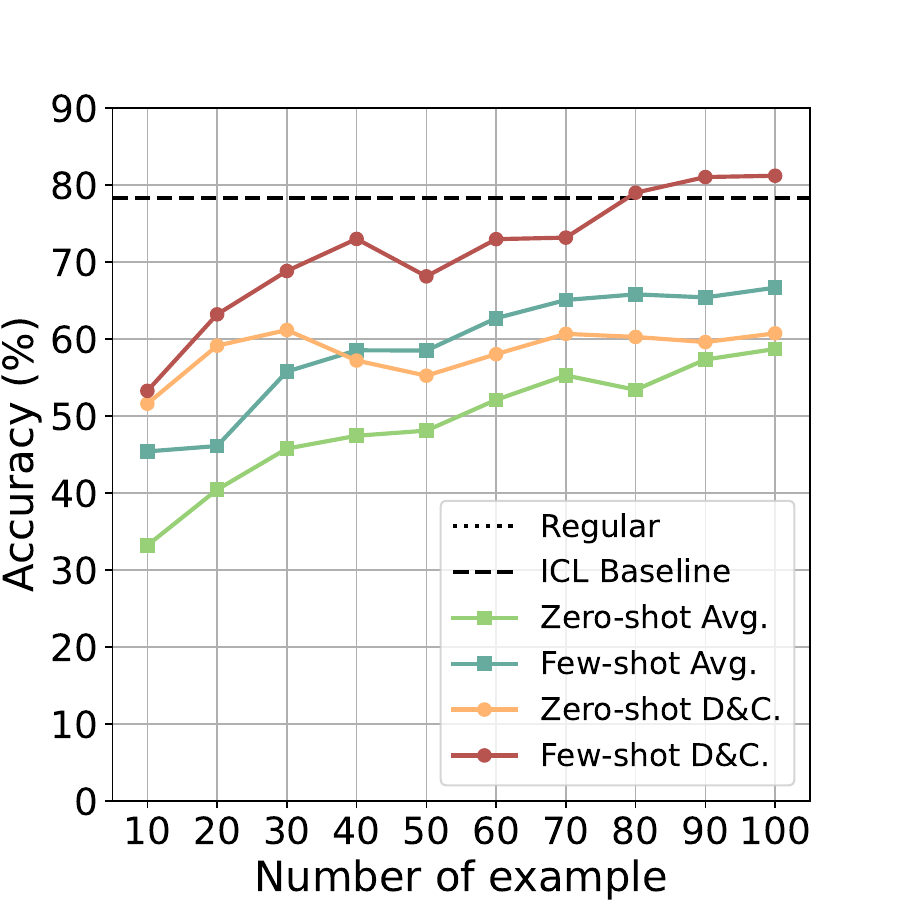}
   \caption{Llama-2 AG News} 
  \end{subfigure} \\
  \begin{subfigure}[b]{0.24\linewidth}
    \includegraphics[width=\linewidth]{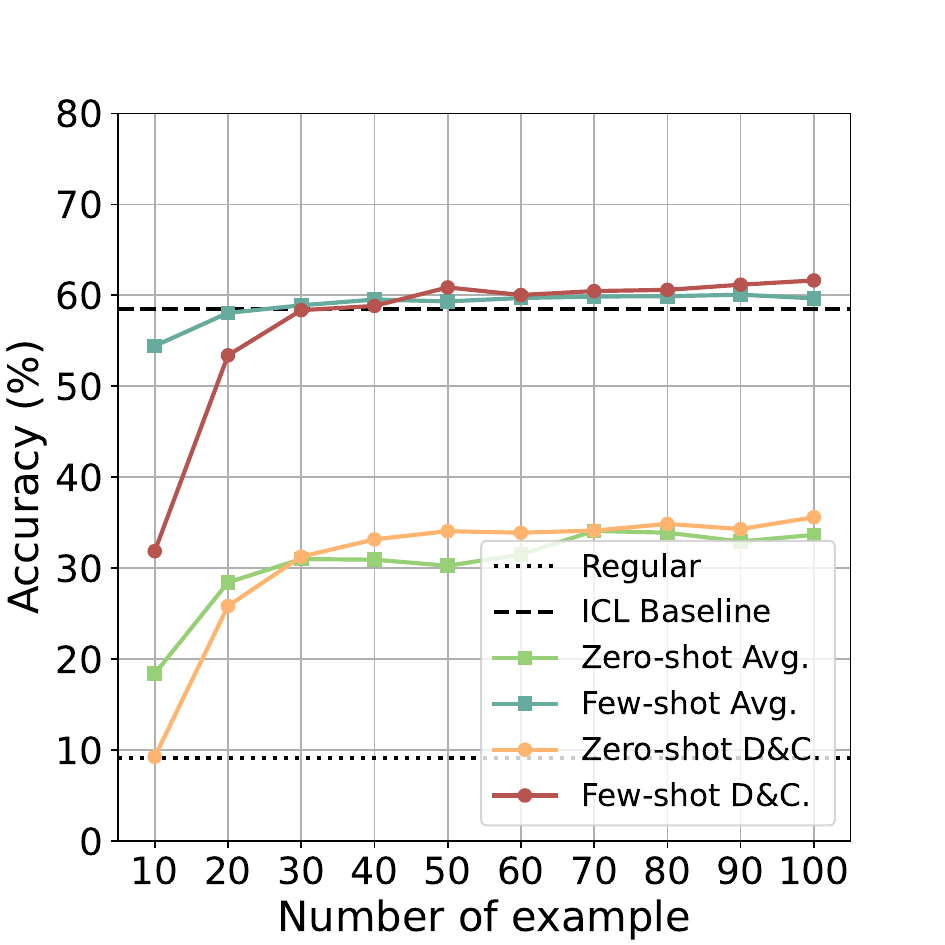}
    \captionsetup{font={scriptsize}}
    \caption{GPT-J Antonym}
    
  \end{subfigure}
  \begin{subfigure}[b]{0.24\linewidth}
    \includegraphics[width=\linewidth]{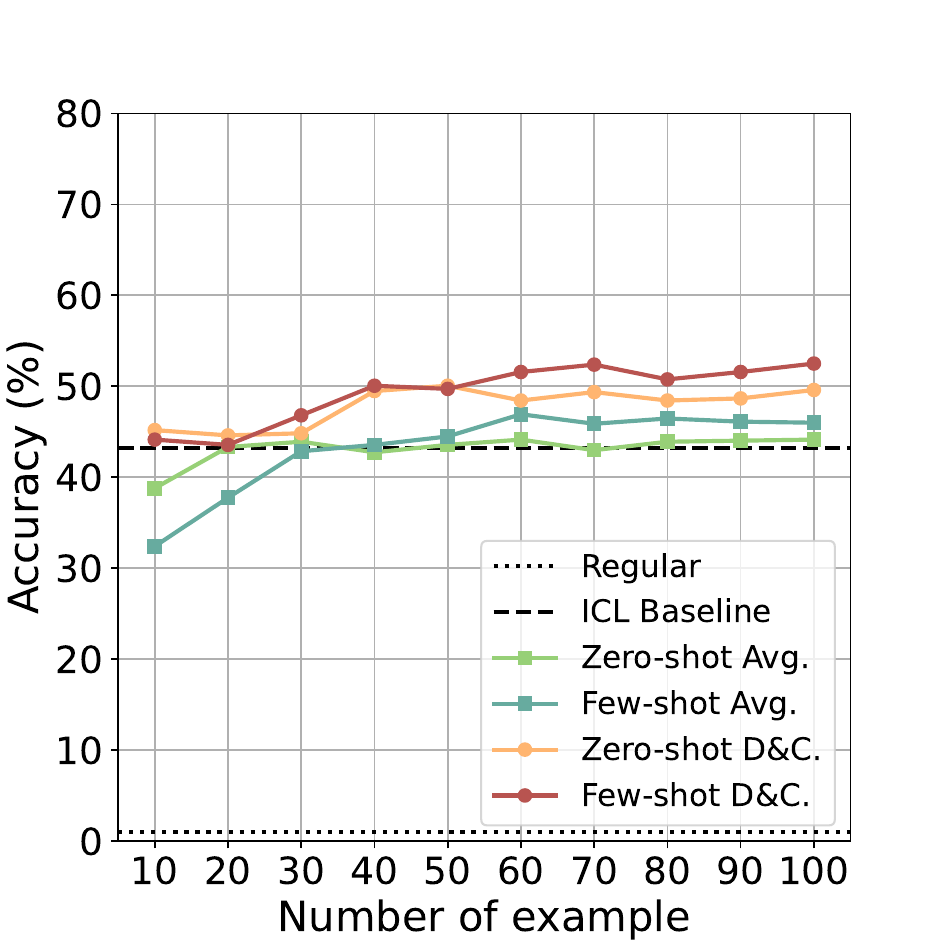}
    \captionsetup{font={scriptsize}}
    \caption{GPT-J Person-Instrument}
    
  \end{subfigure}
  \begin{subfigure}[b]{0.24\linewidth}
    \includegraphics[width=\linewidth]{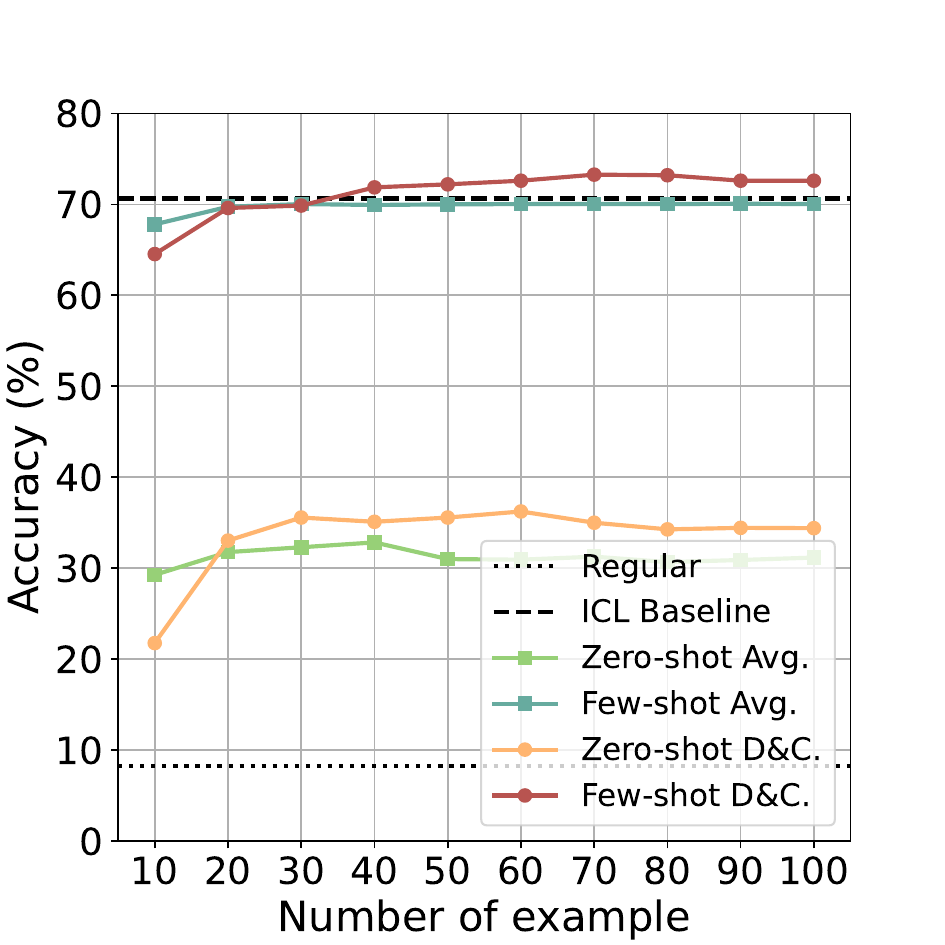}
    \captionsetup{font={scriptsize}}
    \caption{GPT-J English-French}
    
  \end{subfigure}
  \begin{subfigure}[b]{0.24\linewidth}
    \includegraphics[width=\linewidth]{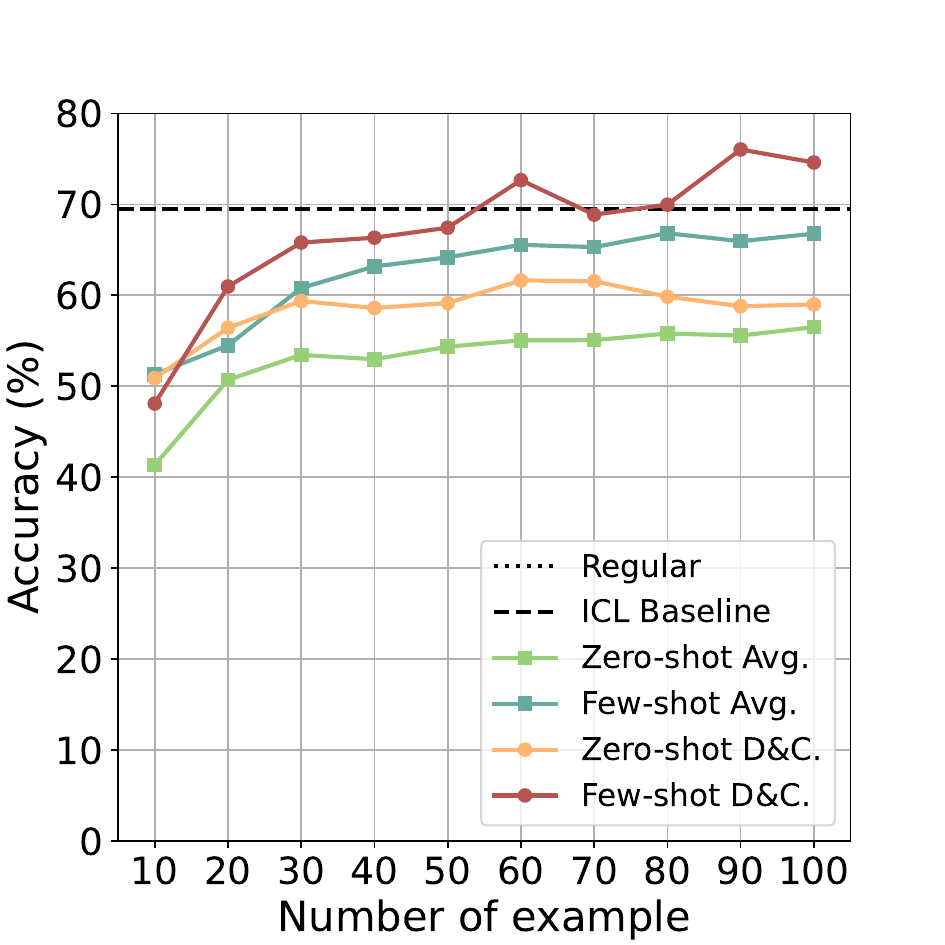}
    \captionsetup{font={scriptsize}}
    \caption{GPT-J AG News} 
   
  \end{subfigure}
  \caption{Performance of aggregation across number of examples. \textit{Avg.} denotes the average aggregation baseline and \textit{D\&C.} denotes the divide-and-conquer aggregation. The \textbf{X} axis represents the number of examples, and the \textbf{Y} axis represents the accuracy.}
  \vspace{-4mm}
  \label{fig:aggregation}
\end{figure*}

\subsection{Inner Optimization(RQ1)}

As shown in Table~\ref{tab:main result}, the performance of our inner optimized  state vector has a significant improvement comparing the task vector and function vector in both zero-shot and few-shot settings. 
Our state vector  with inner optimization. In the zero-shot setting, the inner optimization shows an average improvement of 10.2\% on Llama-2 and 5.9\% on GPT-J across six datasets. 
In the few-shot setting, the inner optimization also achieves a 1.2\% improvement on Llama-2 and 1.7\% on GPT-J. The improvement demonstrates the effectiveness of inner optimization. However, although state vector (inn.) outperforms task vector, its few-shot performance on some datasets is inferior to the ICL baseline. We attribute this primarily to the introduction of query information from examples.
While inner optimization enhances task-relevant information for the state vector, it also introduces noise of other dummy queries, hindering the model's ability to focus on the current predictive query, thereby reducing performance.
In addition to the performance improvements, our inner optimization approach also effectively alleviates the phenomenon of high variance in the original task vector in the zero-shot setting.
In practical use, the performance of the task vector is influenced by demonstrations and dummy queries, leading to weaker robustness. 
Our proposed inner optimization approach effectively mitigates this issue, similarly motivated as the model averaging method, thereby enhancing the robustness of the state vector.

\subsection{Momentum Optimization (RQ2)}


As depicted in Table~\ref{tab:main result}, building upon the inner optimized  state vector, our proposed momentum optimization algorithm further enhances the effectiveness of the state vector, achieving the best performance on average in all settings.
In the zero-shot setting, the momentum optimization boosts the performance of the inner-optimized state vector with an average increase of 1.3\% on Llama-2 and 2.4\% on GPT-J.
In the few-shot setting, state vector with momentum optimization achieves a 0.8\% average increase on Llama-2 and 1.0\% on GPT-J. This reveals the effectiveness of our momentum optimization.
With the combination of inner optimization and momentum optimization, our state vector (mom.) surpasses the original variant, showcasing a remarkable improvement of 11.5\% for Llama-2 and 8.3\% for GPT-J in the zero-shot setting. In the few-shot setting, our state vector (mom.) still outperforms the task vector with a 2.0\% improvement for Llama-2 and 2.7\% for GPT-J.
Furthermore, without inputting demonstration during inference, the state vector (mom.) achieves an impressive 90\% ICL performance on Llama-2 and 78\% ICL performance on GPT-J. When compared to ICL with the same examples as the demonstration, state vector (mom.) outperforms ICL in both Llama-2 and GPT-J. These improvements verify the effectiveness of our progressive optimization strategy. 
Note that applying momentum optimization directly to task vectors does not yield average improvements across tasks in our preliminary experiment. We speculate that this inconsistency stems from the poor robustness of the task vectors, which hinders the stable optimization by momentum optimization and leads to poor performance in some tasks.

\subsection{Divide-and-Conquer Aggregation (RQ3)}

In this experiment, we explore the performance of D\&C state vector aggregation across varying numbers of examples. Besides the regular and ICL baseline mentioned, we introduce average aggregation as a strong baseline. This approach first extracts state vectors from the example group and subsequently employs their mathematical average for aggregation. We compare our D\&C aggregation method with the baseline ranging from 10 to 100 examples across two models. Due to limited computational resources, we were not able to do an exhaustive search over all datasets. Thus, we only present the results for four tasks.

As illustrated in the Figure~\ref{fig:aggregation}, both the D\&C aggregation and average aggregation exhibit similar trends in both few-shot and zero-shot settings. The performance of both aggregation methods initially falls short of the ICL baseline.  However, their performance boosts when examples increase. The initial poor performance can be attributed to the limited number of state vectors. 
Additionally, although the performance of the D\&C aggregation initially falls behind that of the average aggregation, it exhibits a more substantial performance improvement when examples increase, ultimately outperforming average aggregation in the multiple example setting, highlighting the efficiency of D\&C aggregation.

\newfloat{figtab}{htb}{fgtb}
\makeatletter
  \newcommand\figcaption{\def\@captype{figure}\caption}
  \newcommand\tabcaption{\def\@captype{table}\caption}
\makeatother

\begin{figure}
    \begin{minipage}[ht]{.48\linewidth}
        \centering
        \setlength{\tabcolsep}{5pt}
        \begin{tabular}{l l l}
        	\toprule
            \textbf{Method}         &\textbf{Zero-shot}     &\textbf{Few-shot}       \\
            \hline       
            ICL baseline            &0.2\tiny{$\pm$ 0.4}	&71.0\tiny{$\pm$ 10.8}	      \\
            Task vector             &52.9\tiny{$\pm$ 9.4}	&68.5\tiny{$\pm$ 10.5}	     \\ 
            State vector (mom.)     &65.2\tiny{$\pm$ 10.2}	&72.2\tiny{$\pm$ 10.6}		    \\
            State vector (adag.)    &11.7\tiny{$\pm$ 12.0}	&16.1\tiny{$\pm$ 10.2}		    \\
            State vector (rms.)     &0.8\tiny{$\pm$ 0.9}	&1.5\tiny{$\pm$ 1.0}		    \\
            State vector (adam.)    &6.7\tiny{$\pm$ 6.1}	&10.6\tiny{$\pm$ 8.5}		    \\
            \bottomrule
        \end{tabular}
        \tabcaption{Performance comparison of gradient optimization algorithms. The method means the optimization algorithm applied to the $\texttt{opt}(\cdot)$ in Eqn.~\ref{grad_eqn}.}
        \label{tab:gradient optimizer}
    \end{minipage} 
    \hfill
    \begin{minipage}[ht]{.48\linewidth}
    \centering
        \includegraphics[width=\linewidth]{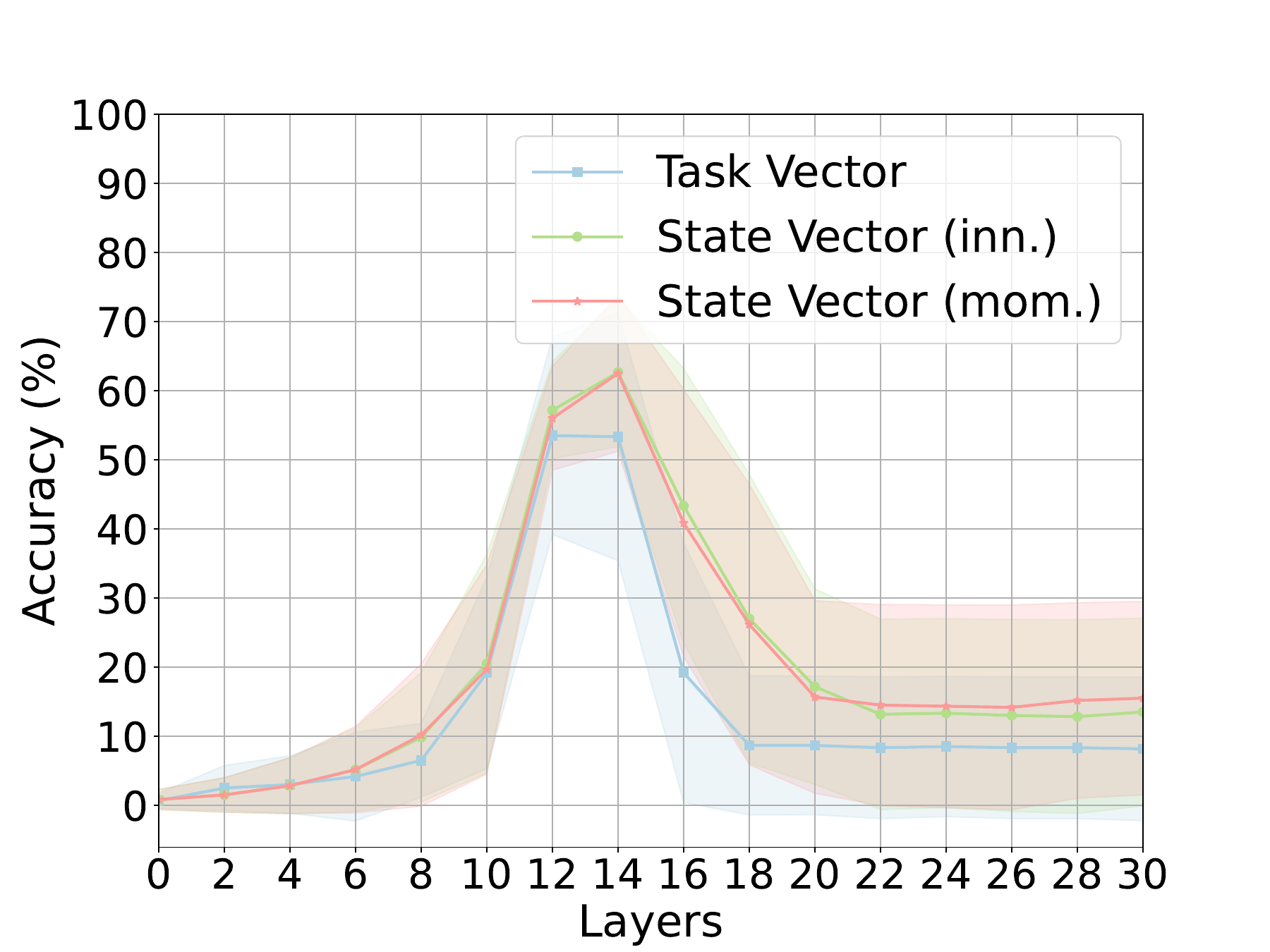}
        \figcaption{Average zero-shot performance across six datasets for each choice of the intermediate layer $L$. The solid line means the average value, while the shaded area indicates the standard deviation.}
        \label{fig:layer select}
    \end{minipage}
\vspace{-4mm}
\end{figure}

\section{Analysis}

\subsection{Ablation with Other Optimization Methods}
\label{sec:GOA}

We present an ablation study to investigate various classical gradient optimization algorithms, aiming to delve deeper into the inner state vector optimization.  We compare the momentum-based gradient optimization algorithm with following additional first-order gradient optimization algorithms: Adagrad (adag.)~\citep{adagrad}, RMSprop (rms.)~\citep{Graves2013GeneratingSW} and Adam(adam.)~\citep{Kingma2014AdamAM}. 
As shown in Table~\ref{tab:gradient optimizer}, we observe a significant decrease in state vector performance with first-order gradient optimization algorithms, unlike with momentum-based optimization.
This outcome indicates a discrepancy between the state vector and updated parameters with gradient descent. It suggests that the current first-order gradient optimization algorithms may not be optimally effective for state vector optimization.
Due to computational constraints, we were not able to do an exhaustive search over all hyper-parameters. 





\subsection{Layer Selection}

We investigate the impact of layer selection on the extraction of state vectors in transformer models. We evaluate the average performance across different datasets in the zero-shot setting, as illustrated in Figure~\ref{fig:layer select}. 
Our results reveal a dual-phase trend:  initially, increasing the number of layers for state vector extraction improves performance, but this improvement  reverses beyond the 14th layer.
We correlate this with the dynamics of ICL function processing in transformers in line with previous works~\citep{DBLP:conf/emnlp/VoitaST19a,DBLP:conf/emnlp/WangLDCZMZS23}. In the initial layers, transformers are primarily engaged in learning and encapsulating the ICL function within state vector, where additional layers enhance the richness of the functional information in the state vector. In contrast, the later layers prioritize applying this learned information for prediction tasks. Here, additional layers tend to introduce noise, especially from predicted labels of dummy queries, which may negatively impact performance.


\begin{figure}[!t]
    \centering
    \begin{subfigure}[b]{0.30\linewidth}
        \includegraphics[width=\linewidth]{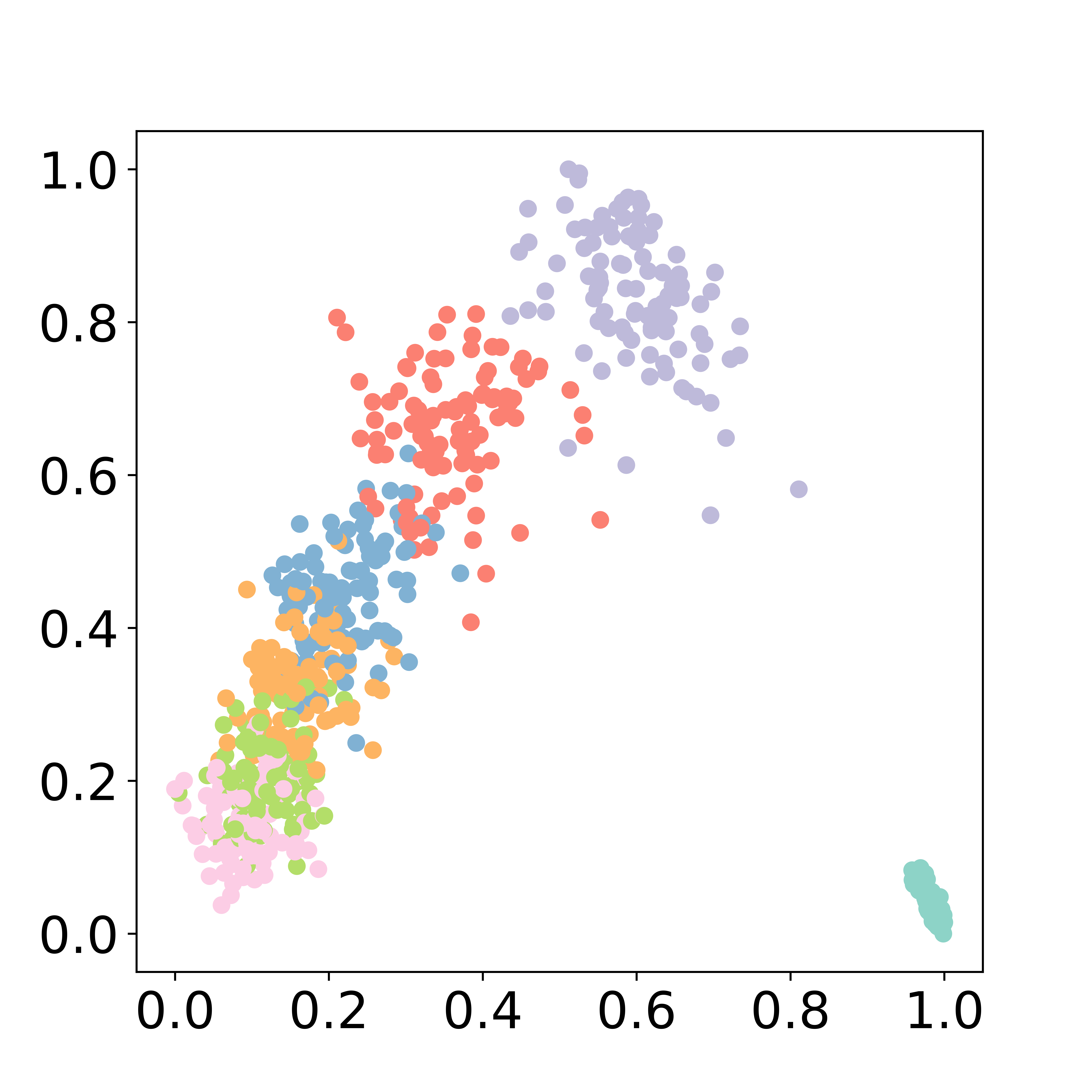}
        \caption{Antonym}
    \end{subfigure} 
    \begin{subfigure}[b]{0.30\linewidth}
        \includegraphics[width=\linewidth]{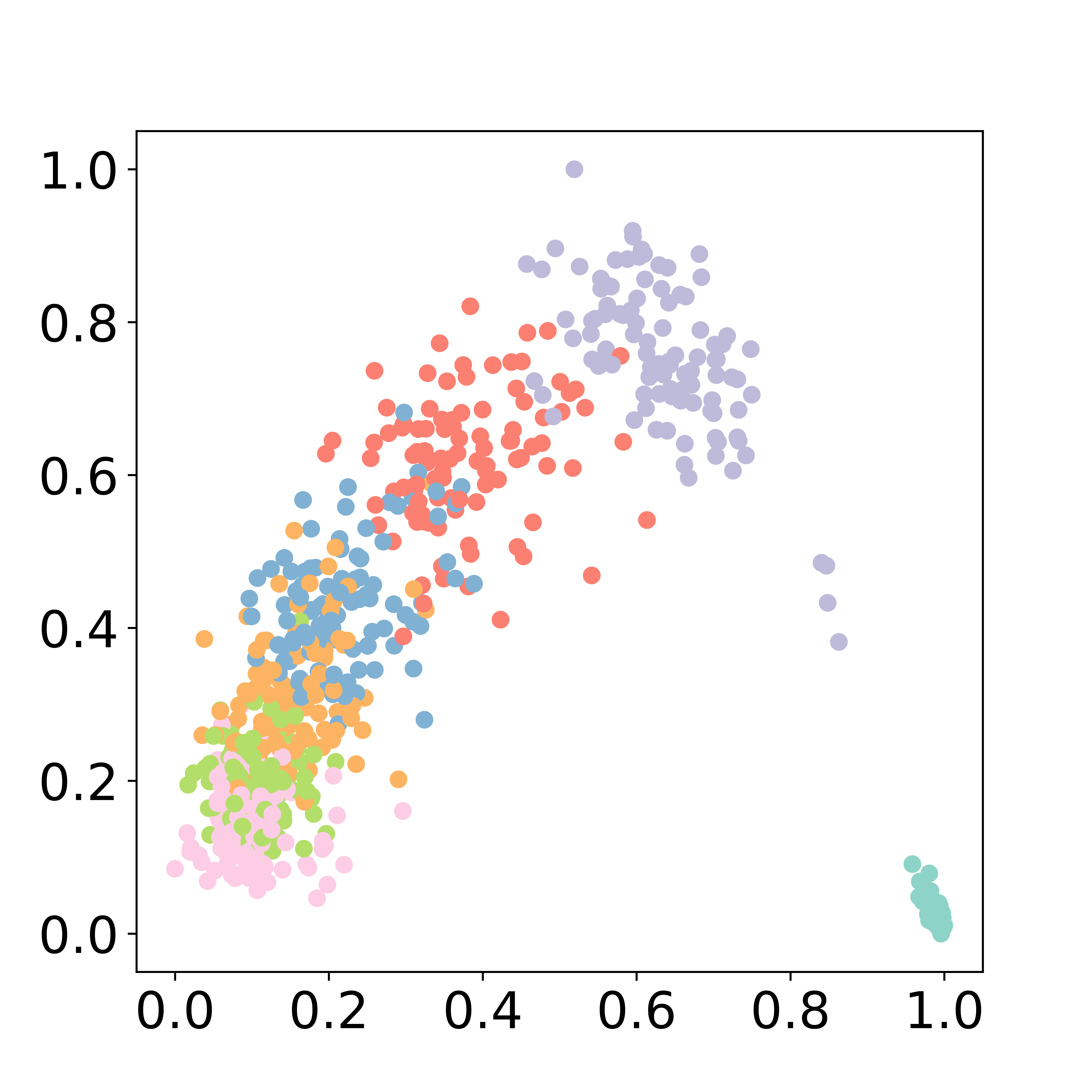}
        \caption{English-French}
    \end{subfigure} 
    \begin{subfigure}[b]{0.30\linewidth}
        \includegraphics[width=\linewidth]{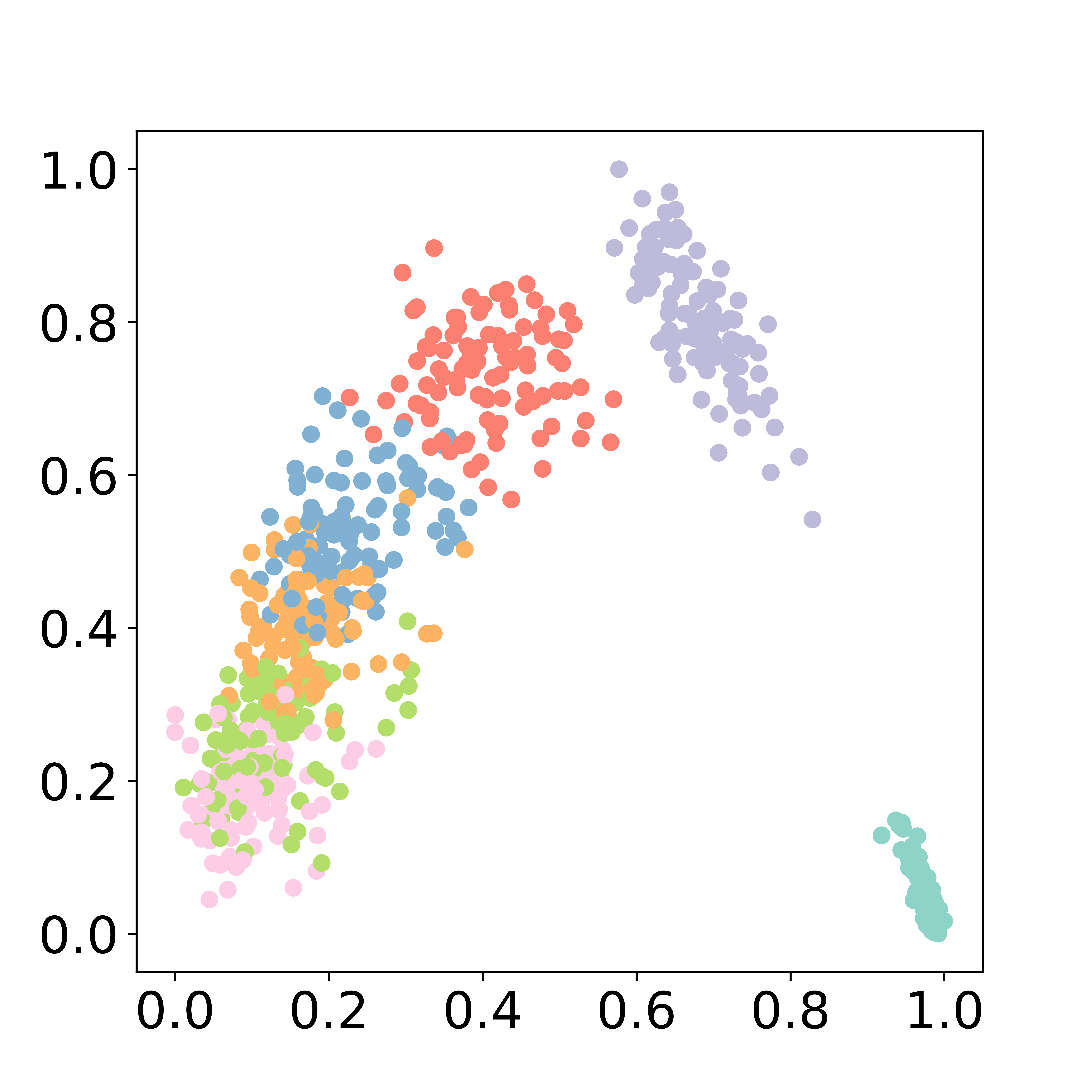}
        \caption{Product-Company}
    \end{subfigure} 
    
    \caption{The 2D PCA visualization of the state vector in the Antonym ,English-French and Product-Company task, where each color represents the state vector corresponding to examples occupying specific positions in the demonstration and the outlier is the first order.}
    \vspace{-4mm}
    \label{fig:qualitative}
\end{figure}

\subsection{Qualitative Study}
\label{sec: Qualitative Study}

We provide the visualization by Principal Component Analysis (PCA) of the original state vector in the Antonym, English-French and Product-Company task. As depicted in Figure~\ref{fig:qualitative}, we have three observations: 
(1) State vectors corresponding to the examples occupying the same position tend to form distinct clusters. This clustering pattern suggests a high degree of similarity among state vectors within each example position, despite different contexts.
(2) A notable separation is evident between the state vectors originating from the first example and other position examples. This demarcation implies that ICL may begin to effectively function with a few examples. 
(3) An interesting trend is observable in the movement of these clusters as the example position increases. This trend may be indicative of an accumulation of task-specific information, where each additional example contributes to a more nuanced understanding of the model. 
These findings suggest a progressive enhancement in the ability of model to internalize and reflect the subtleties of the task at hand.
Moreover, these observations  reflect the efficacy of momentum optimization to leverage the observed clustering trend. 

\section{Conclusion}
In this paper, we reveal that ICL compressed vector can be viewed as parameters trained through gradient descent on the demonstrations. Then, we introduce the concept of state vector coupled with two optimization methods to enhance the capability of ICL and conduct comprehensive experiments across two popular LLMs and multiple tasks to support our claim. Furthermore, our approach demonstrates the ability to compress context while maintaining lower variance. In the future, we aim to extend our methods to more complex ICL scenarios and apply them to larger LLMs and call for more nuanced and realistic studies of ICL.


\bibliographystyle{plainnat}
\bibliography{neurips_2024}

\appendix
\newpage

\section{Implementation Details}
\label{sec:implement details}

In this paper, we use random sampling to create subsets for each dataset. Each subset consists of 10 instances for demonstrations and one instance for a dummy query since we employ a 10-shot as the default ICL setting. The remaining instances are split into test and development sets with a 7:3 ratio. For experiments with multiple examples, we sample 100 instances instead of 10.
We use ``$\rightarrow$'' as the separate token similar to previous works. We tried other tokens but no significant difference. All the experiments are reported over 5 random seeds.
The inference mechanism with state vector we describe in \S\ref{sec:Basic Inference} has a key hyper-parameter (i.e.the layer $L$). Previous studies~\citep{Hendel2023InContextLC} have shown that the choice of $L$ has an influence on performance. We find the best layer for different tasks via the accuracy of the development set.
For the inner optimization in \S\ref{sec:Average Optimize}, we choose the last seven state vectors to optimize. This is because the early state vectors yield subpar performance, primarily due to limitations in the available examples.
For the momentum optimization, we choose 0.5 as the retention rate for historical momentum from the options of 0.25, 0.5 and 0.75.
We run all the experiments on a single NVIDIA A100 80G GPUs. Each of our experiments consumes between 10 minutes to 8 hours of GPU time, depending on the dataset.

\section{More Details about Baseline}

In this section, we present an in-depth and comprehensive analysis of two baselines (i.e. task vector~\citep{Hendel2023InContextLC} and function vector~\citep{Todd2023FunctionVI}). Furthermore, we offer a more nuanced comparison with our proposed state vector, highlighting the distinct differences and advantages of our approach. 

The task vector is designed to extract the ICL function from a specific layer's hidden state within the transformer model. This is achieved by directly replacing the corresponding hidden state during inference for intervention. On the other hand, \cite{Todd2023FunctionVI} first extracts the ICL function from the output activations across all attention heads in all transformer layers. These activations are then prioritized based on their causal effect, quantified by the variance in the model's output space with or without individual activation interventions. The mathematical average of the top 10 causal effect activations is the function vector, which is subsequently added to the hidden state of a specific layer during the inference stage.

In contrast to these methods, our approach for state vector extraction focuses on procuring the ICL procession state from the output activations of the attention heads within the first $L$ layers. During inference, we replace the corresponding activations with optimized ones. While functionally equivalent to the forward process of the task vector when disregarding state vector optimization (i.e., the vanilla state vector), our approach offers enhanced mechanical explainability. This is attributable to its motivation from the dual form of in-context learning and gradient decay, as explicated in previous work~\citep{ICL_Gradient_Descent_as_Meta_Optimizers,Natan2023IncontextLA}. Furthermore, inspired by the dual form, we focus on the further optimization process. On the other hand, unlike the function vector which extracts activations based on the causal effects resulting from individual interventions, our method is rooted in the underlying mechanisms of ICL. 
This strategy not only improves mechanical explainability  but also demonstrates greater performance as evidenced by extensive experiments. Experiments also show notably poor performance of the function vector on certain knowledge-based datasets, such as Person-Occupation.

\section{More Details about Datasets}

Here, we describe in detail the tasks that we use to evaluate the state vectors.

\begin{itemize}
\item \textbf{Antonym}~\citep{Nguyen2017DistinguishingAA} contains 2398 word pairs that are antonyms of each other (e.g. ``massive'' $\rightarrow$ ``tiny''). We apply the dataset processed version from the function vector~\citep{Todd2023FunctionVI}. They filter the word pairs where both words can be tokenized as a single token.
\item \textbf{Capitalize}~\citep{Todd2023FunctionVI} contains 813 word pairs that capitalize the first letter of the given input word (e.g. ``plan'' $\rightarrow$ ``Plan'').
\item \textbf{Present-Past}~\citep{Todd2023FunctionVI} contains 293 word pairs, where simple past tense verbs are output when given simple present tense verbs (e.g. ``adapt'' $\rightarrow$ ``adapted'').
\item \textbf{Singular-Plural}~\citep{Todd2023FunctionVI} contains 205 word pairs, where the plural form of a given singular word (e.g., ``wallet'' $\rightarrow$ ``wallets'').
\item \textbf{English-French}~\citep{Conneau2017WordTW} contains 4698 pairs of words, which consists of a word in English and its translation into French (e.g., ``circle'' $\rightarrow$ ``cercle''). We apply the processed version from the function vector\citep{Todd2023FunctionVI}.
\item \textbf{Country-Capital}~\citep{Todd2023FunctionVI} contains 197 instances, which output  the name of the capital city of the given country (e.g. ``Luanda'' $\rightarrow$ ``Angola'').
\item \textbf{AG News}~\citep{DBLP:conf/nips/ZhangZL15} contains 7600 instances. Each instance contains the news headlines and the first few sentences of an article as input, and output corresponding labels include Business, Science, Sports, and World.
\item \textbf{Person-Sport}~\citep{Hernandez2023LinearityOR} contains 318 instances. Each instance contains the name of a professional athlete and the sport that they play (e.g. ``Hank Aaron'' $\rightarrow$ ``basketball'').
\item \textbf{Person-Instrument}~\citep{Hernandez2023LinearityOR} contains 510 instances. Each instance contains the name of a professional musician and the instrument they play (e.g. ``Tom Fletcher'' $\rightarrow$ ``guitar'').
\item \textbf{Person-Occupation}~\citep{Hernandez2023LinearityOR} contains 821 instances. Each instance contains the name of a well-known individual and their occupation (e.g. ``Tom Fletcher'' $\rightarrow$ ``guitar'').
\item \textbf{Product-Company}~\citep{Hernandez2023LinearityOR} contains 522 instances. Each instance contains the name of a commercial product and the company that sells the product (e.g. ``Tom Fletcher'' $\rightarrow$ ``guitar'').
\item \textbf{Landmark-Country}~\citep{Hernandez2023LinearityOR} contains 836 instances. Each instance contains the name of a landmark and the country in which it is located.
\end{itemize}

\section{Efficiency Analysis}

\begin{figure}[!t]
    \centering
    \includegraphics[width=0.7\linewidth]{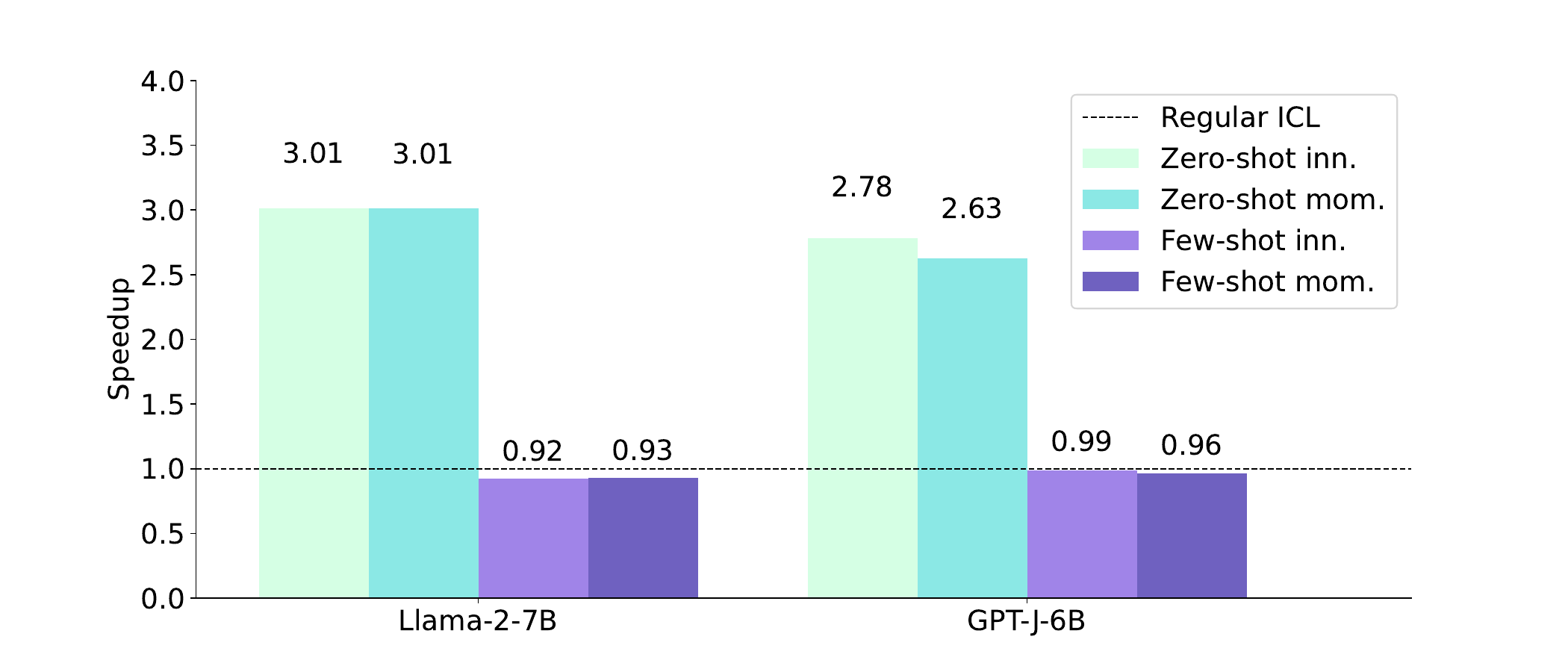}
    \caption{Time efficiency analysis of Llama-2-7B and GPT-J-6B. Inn denotes our state vector with inner optimization. Mom denotes our state vector with momentum optimization}
    \label{fig:efficiency}
\end{figure}

In this section, we present an efficiency analysis of two proposed optimization methods. We evaluate the average inference time using 1000 test data on a single NVIDIA A100 (80G) GPU, covering six main datasets and 10 random seeds per dataset. The results are illustrated in Figure~\ref{fig:efficiency}.
In the zero-shot setting, we compress the ICL function into the state vector which eliminates the need to concatenate demonstrations during inference. As shown in the Figure~\ref{fig:efficiency}, the proposed inner optimization and momentum optimization, which, while tripling the inference speed, achieve 89\% of the regular ICL performance on Llama-2-7B and 78\% on GPT-J-6B (see Table 1 in the paper).
In the few-shot setting, the proposed inner optimization and momentum optimization achieve better results than standard ICL at the cost of a minimal loss in inference speed (e.g., 99\% and 96\%). 
Moreover, our method is orthogonal to attention speedup techniques, such as flash attention~\citep{flash_attention} and page attention~\citep{vllm}. Therefore, our approach can also benefit from the achievements of these works and achieve further efficiency improvement. We leave the exploration of alternative enhancement as future work.

\begin{table}[t]
\centering

	\begin{tabular}{l l l}
		\toprule
        \textbf{Prompt}         &\textbf{Llama-2}     &\textbf{+SV}       \\
        \hline       
        What instrument did X play? &8.7\tiny{$\pm$ 0.7}	&67.3\tiny{$\pm$ 2.8 } \\
        Can you tell me which musical instrument was played by X?	
        &25.1\tiny{$\pm$ 0.7}	&69.0\tiny{$\pm$ 4.3} \\ 
        What was the primary instrument of X in their music career?
        &17.3\tiny{$\pm$ 1.4}	                  &70.3\tiny{$\pm$ 2.9} \\
        \bottomrule
	\end{tabular}
 \vspace{2mm}
    \caption{Text portability of momentum optimized state vector. The templates are provided with ``X'' replaced by a query word. ``+SV'' denotes adding momentum optimized state vector}
    \vspace{-4mm}
    \label{tab:natural prompt}
\end{table}

\section{Natural Text Completions}

In this study, we evaluate the effectiveness of the momentum optimized state vector on natural text completions. Given a natural text template, we instruct the model to greedily generate 5 tokens with or without intervention in the zero-shot setting. We use exact match accuracy as the metric.
Table~\ref{tab:natural prompt} shows the result of natural text completions on Llama-2. The performance boosts observed with the momentum-optimized state vector on the separate tokens indicate that it can guide the model to generate answers correctly. We include more examples of natural text completions in the Appendix.

\section{Case Study}

\begin{table}[t]
\centering
	\begin{tabular}{l l}
		\toprule
        \textbf{English-French}  & \\
        \cmidrule(lr){0-0}
        \textbf{Prompt} &What is the meaning of biography?      \\
        \hline       
        Llama-2       &A written account of someone’s life.   \\
        \quad + state vector   &It is biographie.   \\
        \midrule  
        \textbf{Antonym}  & \\
        \cmidrule(lr){0-0}
        \textbf{Prompt}	&When I think of upright, I think of \\ 
        \hline
         Llama-2       &I think of a person who is standing up \\                   &for what they believe in. \\
        \quad + state vector   &I think of down.   \\
        \bottomrule
	\end{tabular}
 \vspace{2mm}
    \caption{Natural prompt cases with momentum optimized state vector on Antonym task and English-French task.}
    \label{tab:case study}
\end{table}

In this section, we present a case study shown in Table~\ref{tab:case study}, to demonstrate the efficacy of the momentum-optimized state vector in natural text completions. Consider the query: ``What is the meaning of biography?'', The vanilla Llama-2 model would directly answer this question. However, when influenced by an English-French state vector, Llama-2 changes its response, translating the question into French instead. Similarly, when presented with the sentence ``When I think of upright, I think of''. Influenced by an Antonym state vector, Llama-2 completes the sentence with an anonymous pattern. These instances exemplify the model learning the ICL function stored in the momentum optimized state vector, enabling it to generate context relevant to the specified task.

\section{Full Result}
\begin{table*}[t]
\centering
\scriptsize
{\renewcommand{\arraystretch}{1.1}%
\setlength{\tabcolsep}{4pt}
\resizebox{\linewidth}{!}{ 
 \begin{tabular}{c|l|l|ccc ccc|c}
  \hline
        \textbf{Model} &\multicolumn{2}{c|}{\textbf{Method}} &\textbf{Capitalize}  &\textbf{Country-Capital} &\textbf{Present-Past}  &\textbf{Singular-Plural}  &\textbf{Person-Sport} &\textbf{AG News} &\textbf{Average (All)}\\
        \hline       
        \multirow{10}{*}{Llama-2}
            &\multirow{5}{*}{Zero-shot}       
                &Regular                                &0.0\tiny{$\pm$ 0.0}                  &0.0\tiny{$\pm$ 0.0}            &0.0\tiny{$\pm$ 0.0}                &0.0\tiny{$\pm$ 0.0}            &0.0\tiny{$\pm$ 0.0}              &0.0\tiny{$\pm$ 0.0}            &0.2    \\
            &   &Function vector                        &98.6\tiny{$\pm$ 0.4}                 &67.4\tiny{$\pm$ 20.7}          &80.2\tiny{$\pm$ 4.5}               &94.2\tiny{$\pm$ 0.6}           &1.4\tiny{$\pm$ 0.5}              &\uline{57.7}\tiny{$\pm$ 0.9}   &44.7   \\
            &   &Task vector                            &92.9\tiny{$\pm$6.5}                  &92.8\tiny{$\pm$2.8}            &95.2\tiny{$\pm$1.7}                &95.3\tiny{$\pm$1.9}            &86.9\tiny{$\pm$4.5}              &47.8\tiny{$\pm$1.3}            &69.4   \\
            &   &State vector (inn.)                    &\uline{99.6}\tiny{$\pm$0.4}          &94.0\tiny{$\pm$1.3}            &\uline{96.5}\tiny{$\pm$1.2}        &\uline{97.1}\tiny{$\pm$1.0}    &\uline{89.7}\tiny{$\pm$3.2}      &52.0\tiny{$\pm$5.5}            &76.0   \\
            &   &State vector (mom.)                    &99.1\tiny{$\pm$0.3}                  &\uline{94.5}\tiny{$\pm$0.7}    &\uline{96.5}\tiny{$\pm$0.7}        &96.6\tiny{$\pm$1.0}            &88.1\tiny{$\pm$2.6}              &50.0\tiny{$\pm$8.3}            &\uline{76.3}   \\
            \cline{2-10}
            &\multirow{5}{*}{Few-shot}     
                &ICL baseline                           &\textbf{99.9}\tiny{$\pm$0.1}         &\textbf{95.2}\tiny{$\pm$1.0}   &\textbf{98.3}\tiny{$\pm$0.6}       &98.5\tiny{$\pm$0.1}            &94.8\tiny{$\pm$0.2}              &76.0\tiny{$\pm$5.7}            &83.1         \\
            &   &Function vector                        &99.7\tiny{$\pm$ 0.1}                 &82.2\tiny{$\pm$ 3.8}           &94.6\tiny{$\pm$ 1.7}               &97.3\tiny{$\pm$ 0.7}           &88.4\tiny{$\pm$ 1.9}             &\textbf{80.7}\tiny{$\pm$4.6}   &78.1       \\
            &   &Task vector                            &98.0\tiny{$\pm$1.0}                  &92.9\tiny{$\pm$3.4}            &98.2\tiny{$\pm$0.5}                &98.5\tiny{$\pm$1.3}            &95.4\tiny{$\pm$0.4}              &64.3\tiny{$\pm$8.4}            &81.5       \\
            &   &State vector (inn.)                    &99.7\tiny{$\pm$0.1}                  &94.4\tiny{$\pm$1.3}            &\textbf{98.3}\tiny{$\pm$0.6}       &98.5\tiny{$\pm$0.4}            &95.2\tiny{$\pm$0.2}              &76.0\tiny{$\pm$8.5}            &83.3         \\
            &   &State vector (mom.)                    &99.3\tiny{$\pm$0.1}                  &94.9\tiny{$\pm$0.7}            &\textbf{98.3}\tiny{$\pm$0.6}       &\textbf{98.8}\tiny{$\pm$0.3}   &\textbf{95.7}\tiny{$\pm$0.2}     &76.3\tiny{$\pm$5.9}            &\textbf{83.8}        \\
            \noalign{\hrule height 1pt}
        \multirow{10}{*}{GPT-J} 
            &\multirow{5}{*}{Zero-shot}    
                &Regular                                &0.3\tiny{$\pm$ 0.1}                  &1.8\tiny{$\pm$ 1.7}            &19.4\tiny{$\pm$ 2.1}               &22.7\tiny{$\pm$ 2.9}           &0.0\tiny{$\pm$ 0.0}              &0.0\tiny{$\pm$ 0.0}            &5.2    \\
            &   &Function vector                        &\uline{66.3}\tiny{$\pm$ 8.4}         &\uline{57.0}\tiny{$\pm$ 9.9}   &\uline{63.1}\tiny{$\pm$ 2.1}       &\uline{69.3}\tiny{$\pm$ 2.1}   &0.8\tiny{$\pm$ 1.1}              &46.4\tiny{$\pm$ 4.5}           &37.4   \\
            &   &Task vector                            &51.0\tiny{$\pm$4.7}                  &31.6\tiny{$\pm$4.8}            &37.0\tiny{$\pm$5.3}                &61.6\tiny{$\pm$1.2}            &46.4\tiny{$\pm$4.0}              &55.0\tiny{$\pm$3.7}            &41.4   \\
            &   &State vector (inn.)                    &58.2\tiny{$\pm$1.3}                  &45.5\tiny{$\pm$8.3}            &47.3\tiny{$\pm$2.0}                &61.9\tiny{$\pm$0.7}            &\uline{51.7}\tiny{$\pm$1.8}      &59.7\tiny{$\pm$5.4}            &47.8\\ 
            &   &State vector (mom.)                    &58.6\tiny{$\pm$0.8}                  &52.9\tiny{$\pm$6.1}            &45.9\tiny{$\pm$0.2}                &62.5\tiny{$\pm$0.7}            &51.4\tiny{$\pm$1.4}              &\uline{61.3}\tiny{$\pm$4.8}    &\uline{49.7} \\
            \cline{2-10}
            &\multirow{5}{*}{Few-shot}     
                &ICL regular                            &99.3\tiny{$\pm$0.3}                  &88.2\tiny{$\pm$3.4}            &96.9\tiny{$\pm$0.9}                &99.3\tiny{$\pm$0.5}            &82.4\tiny{$\pm$3.5}              &76.3\tiny{$\pm$1.7}            &73.1   \\
            &   &Function vector                        &98.6\tiny{$\pm$ 0.6}                 &78.6\tiny{$\pm$ 5.1}           &90.8\tiny{$\pm$ 1.3}               &95.9\tiny{$\pm$ 0.9}           &81.6\tiny{$\pm$ 1.4}             &72.7\tiny{$\pm$3.2}            &70.6      \\  
            &   &Task vector                            &99.3\tiny{$\pm$0.3}                  &89.8\tiny{$\pm$2.8}            &97.3\tiny{$\pm$1.0}                &99.3\tiny{$\pm$0.5}            &83.3\tiny{$\pm$3.6}              &63.3\tiny{$\pm$8.7}            &71.7   \\
            &   &State vector (inn.)                    &\textbf{99.4}\tiny{$\pm$0.3}         &89.2\tiny{$\pm$3.6}            &97.3\tiny{$\pm$0.8}                &99.3\tiny{$\pm$0.5}            &\textbf{83.8}\tiny{$\pm$3.5}     &75.7\tiny{$\pm$1.2}            &73.6   \\
            &   &State vector (mom.)                    &\textbf{99.4}\tiny{$\pm$0.2}         &\textbf{90.1}\tiny{$\pm$3.5}   &\textbf{97.6}\tiny{$\pm$0.9}       &\textbf{99.4}\tiny{$\pm$0.3}   &83.7\tiny{$\pm$3.0}              &\textbf{78.0}\tiny{$\pm$2.2}   &\textbf{74.4}   \\
        \hline
    \end{tabular}
    }
 }
    \caption{Performance of state vector optimization across other six tasks and average performance of all task. The best results in the zero shot setting are in \uline{underline} and the best results in the few shot setting are in \textbf{bold}. The result of basic state vector is mathematically equivalent to task vector.}
    \vspace{-4mm}
    \label{tab:main result appendix}
\end{table*}

In this section, we provide the additional result with llama-2-7B GPT-J model. We first present the main result of optimization on the other six tasks except the main result, and the average performance across all tasks. As shown in Table~\ref{tab:main result appendix}, our inner optimization and momentum optimization effectively enhance the state vector.

\begin{figure*}[hpt]
\centering
\begin{subfigure}[b]{0.24\textwidth}
    \includegraphics[width=\linewidth]{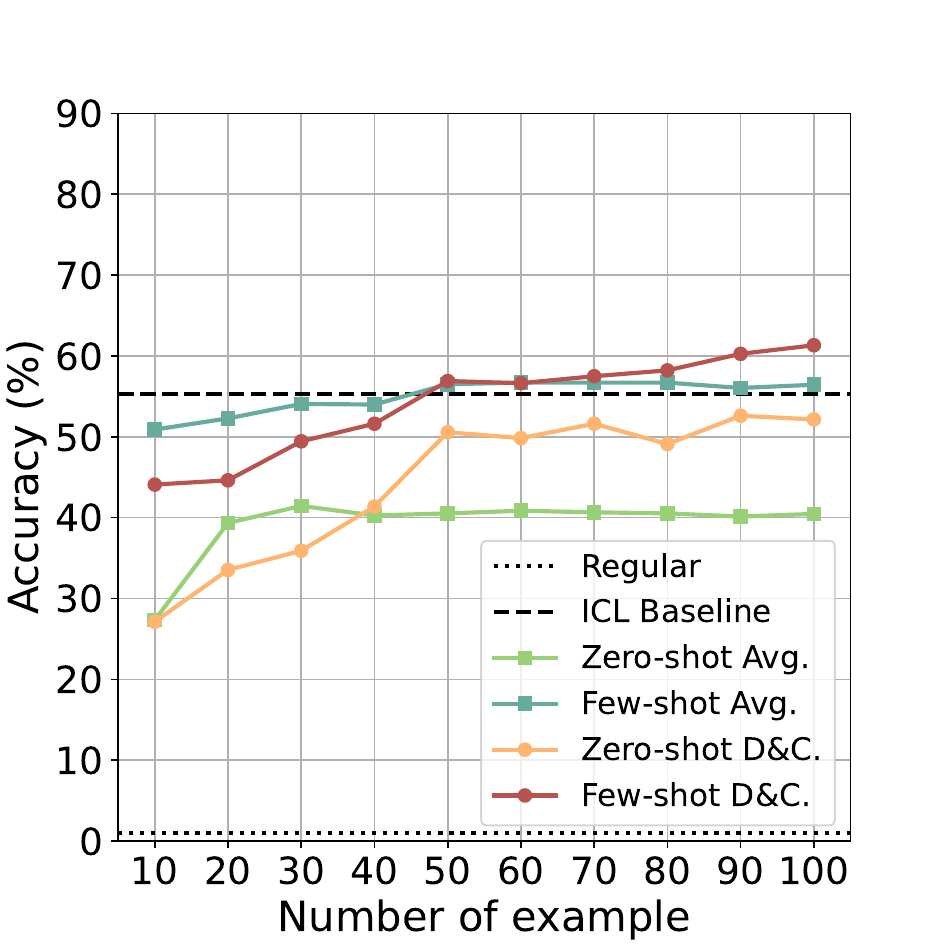}
    \captionsetup{font={scriptsize}}
    \caption{Llama-2 Person-Occupation}
  \end{subfigure}
  \begin{subfigure}[b]{0.24\textwidth}
    \includegraphics[width=\linewidth]{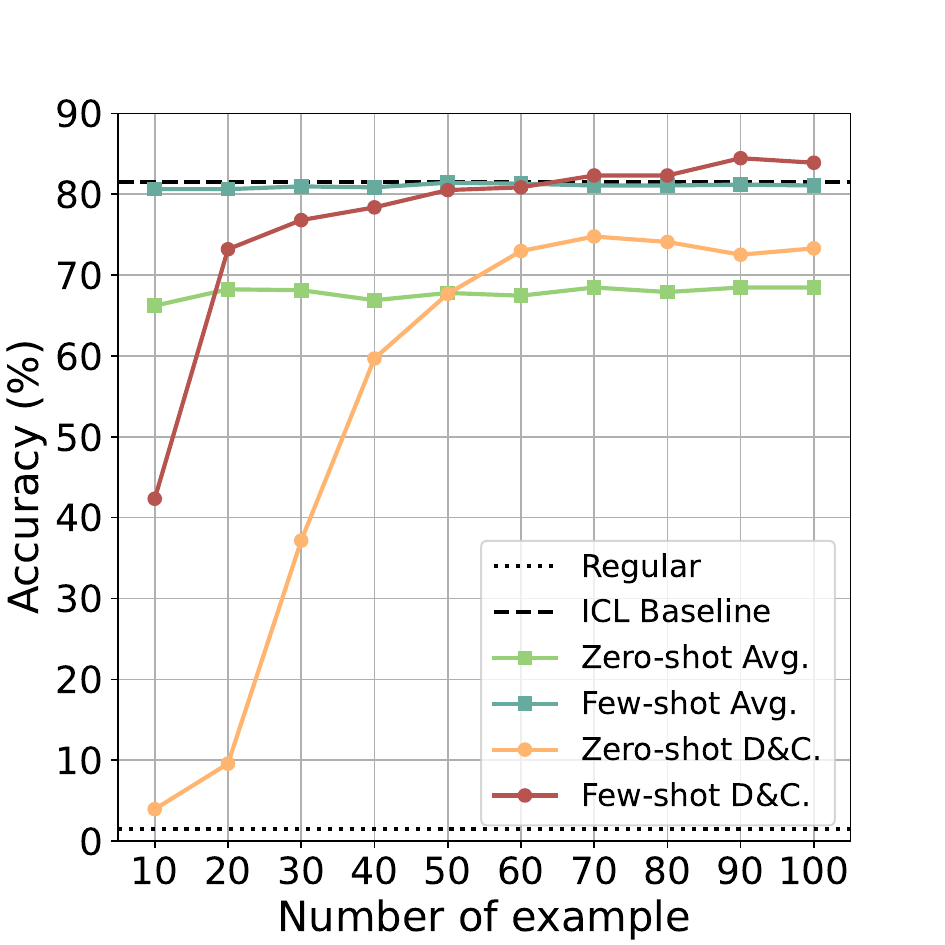}
    \captionsetup{font={scriptsize}}
    \caption{Llama-2 Product-Company}
  \end{subfigure}
  \begin{subfigure}[b]{0.24\textwidth}
    \includegraphics[width=\linewidth]{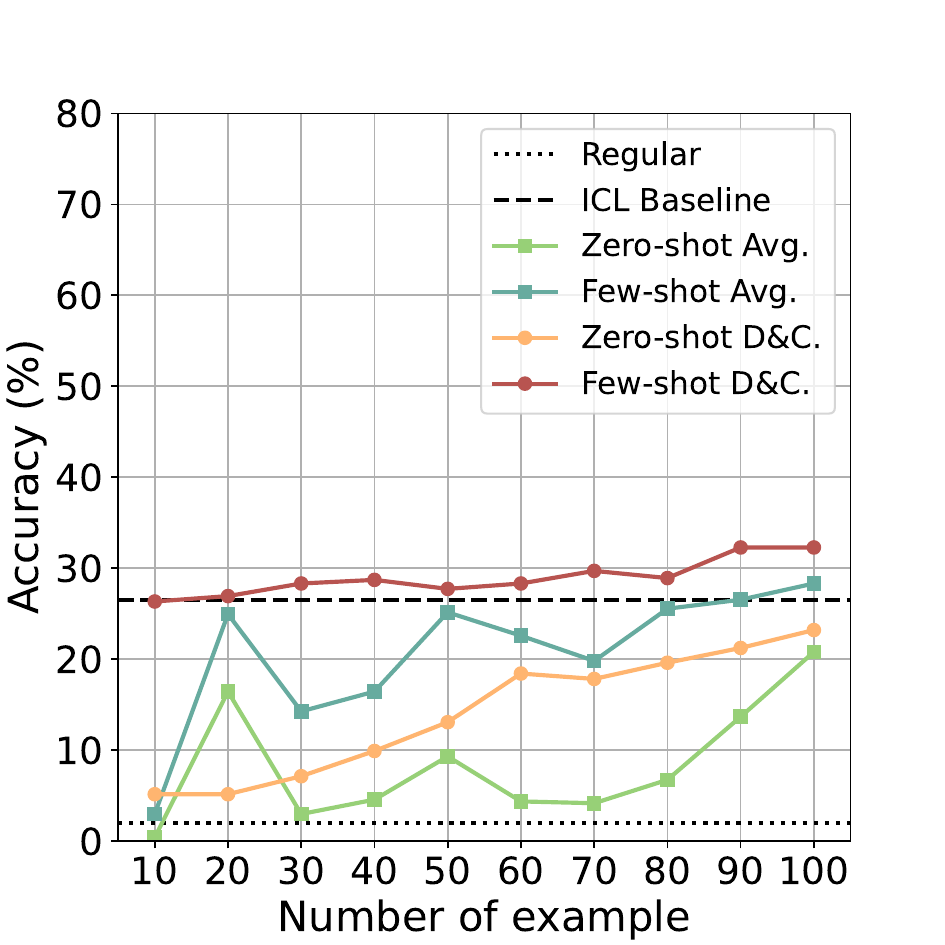}
    \captionsetup{font={scriptsize}}
    \caption{GPT-J Person-Occupation}
    
  \end{subfigure}
  \begin{subfigure}[b]{0.24\textwidth}
    \includegraphics[width=\linewidth]{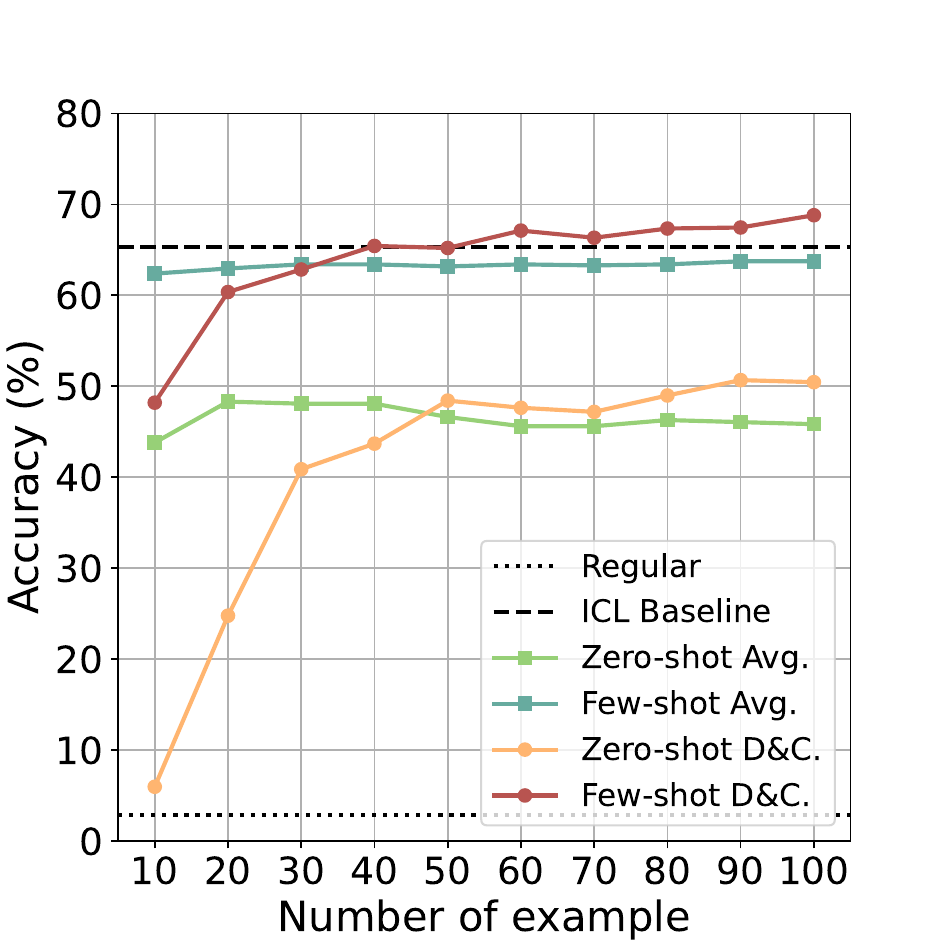}
    \captionsetup{font={scriptsize}}
    \caption{GPT-J Product-Company}
    
  \end{subfigure}
  \caption{Performance of aggregation across number of examples. \textit{Avg.} denotes the average aggregation baseline and \textit{D\&C.} denotes the divide-and-conquer aggregation. The \textbf{X} axis represents the number of examples, and the \textbf{Y} axis represents the accuracy.}
  \label{fig:aggregation_appendix}
\end{figure*}

Moreover, we provide the result of state vector aggregation on two additional datasets. As shown in Figure~\ref{fig:aggregation_appendix}, the trends of both D\$C and average aggregation follow a similar pattern to the main result shown in Figure~\ref{fig:aggregation} as the number of examples increases, illustrating the effectiveness of our aggregation methods.

\section{Result on Larger Model}
\label{sec: larger model}

\begin{table*}[t]
\centering
\scriptsize
{\renewcommand{\arraystretch}{1.1}%
\setlength{\tabcolsep}{4pt}
\resizebox{\linewidth}{!}{ 
 \begin{tabular}{c|l|l|ccc|c}
  \hline
        \textbf{Model} &\multicolumn{2}{c|}{\textbf{Method}} &\textbf{Antonym}  &\textbf{English-French} &\textbf{Person-Instrument}  &\textbf{Average}\\
        \hline       
        \multirow{10}{*}{Llama-2-13B}
            &\multirow{5}{*}{Zero-shot}       
                &Regular                &1.2\tiny{$\pm$ 0.7}            &0.2\tiny{$\pm$ 0.2}        &0.0\tiny{$\pm$ 0.0}            &0.5    \\
            &   &Function vector        &47.1\tiny{$\pm$ 1.6}           &23.2\tiny{$\pm$4.3}        &0.1\tiny{$\pm$ 0.1}            &23.5   \\
            &   &Task vector            &46.0\tiny{$\pm$ 2.4}           &43.1\tiny{$\pm$7.2}        &58.2\tiny{$\pm$6.3}            &49.1   \\
            &   &State vector (inn.)    &47.0\tiny{$\pm$ 1.2}           &50.5\tiny{$\pm$1.9}        &66.6\tiny{$\pm$3.1}            &54.7   \\
            &   &State vector (mom.)    &\uline{47.9\tiny{$\pm$ 1.1}}   &\uline{55.9\tiny{$\pm$3.4}}    &\uline{68.5\tiny{$\pm$2.0}}    &\uline{57.4}   \\
            \cline{2-7}
            &\multirow{5}{*}{Few-shot}     
                &ICL baseline           &\textbf{67.0\tiny{$\pm$ 0.1}}  &74.5\tiny{$\pm$1.3}                &75.0\tiny{$\pm$0.2}            &72.2   \\
            &   &Function vector        &65.7\tiny{$\pm$ 1.7}           &75.2\tiny{$\pm$2.6}                &72.2\tiny{$\pm$0.4}            &71.3   \\
            &   &Task vector            &64.8\tiny{$\pm$ 1.2}           &70.5\tiny{$\pm$3.5}                &70.6\tiny{$\pm$3.1}            &68.6   \\
            &   &State vector (inn.)    &65.5\tiny{$\pm$ 0.8}           &\textbf{75.8\tiny{$\pm$1.6}}       &77.0\tiny{$\pm$1.3}            &72.8   \\
            &   &State vector (mom.)    &65.9\tiny{$\pm$ 0.7}           &75.6\tiny{$\pm$0.4}                &\textbf{78.6\tiny{$\pm$1.1}}   &\textbf{73.4}   \\
        \hline
    \end{tabular}
    }
 }
    \caption{Performance of state vector optimization across three tasks on llama-2-13B. The best results in the zero shot setting are in \uline{underline} and the best results in the few shot setting are in \textbf{bold}. The result of basic state vector is mathematically equivalent to task vector.}
    \vspace{-4mm}
    \label{tab:main result appendix llama13B}
\end{table*}

\begin{figure*}[hpt]
\centering
\begin{subfigure}[b]{0.24\textwidth}
    \includegraphics[width=\linewidth]{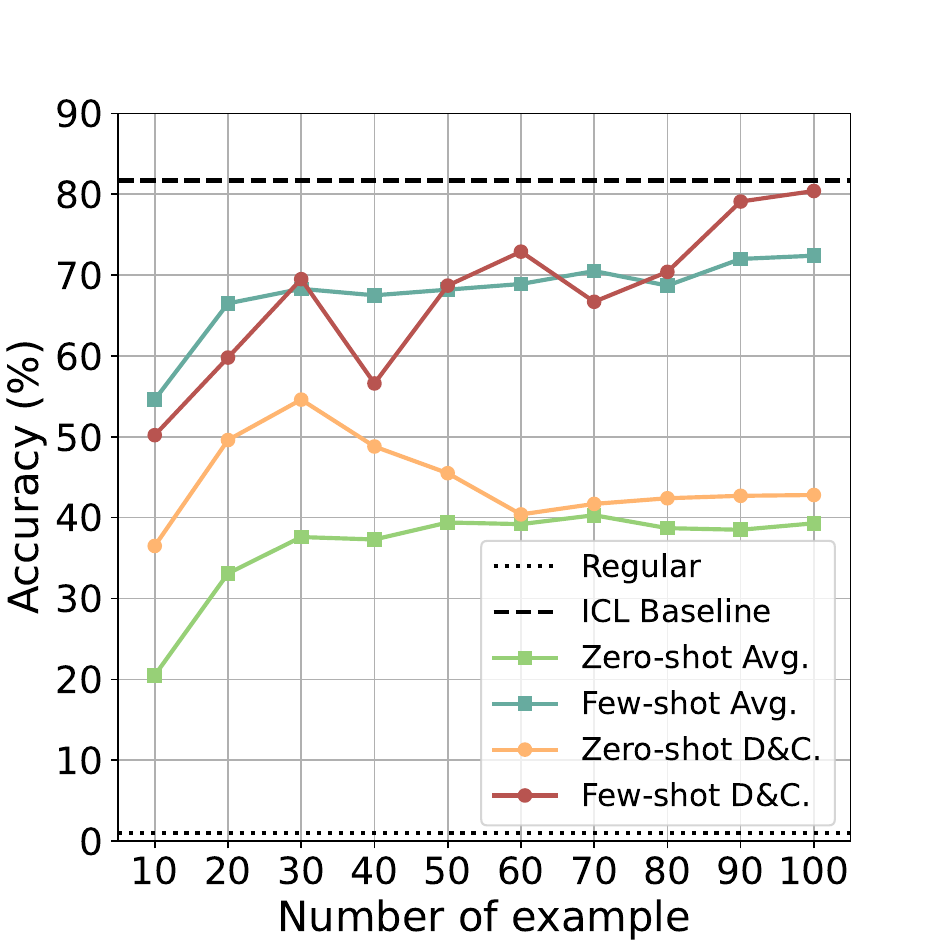}
    \captionsetup{font={scriptsize}}
    \caption{AG News}
  \end{subfigure}
  \begin{subfigure}[b]{0.24\textwidth}
    \includegraphics[width=\linewidth]{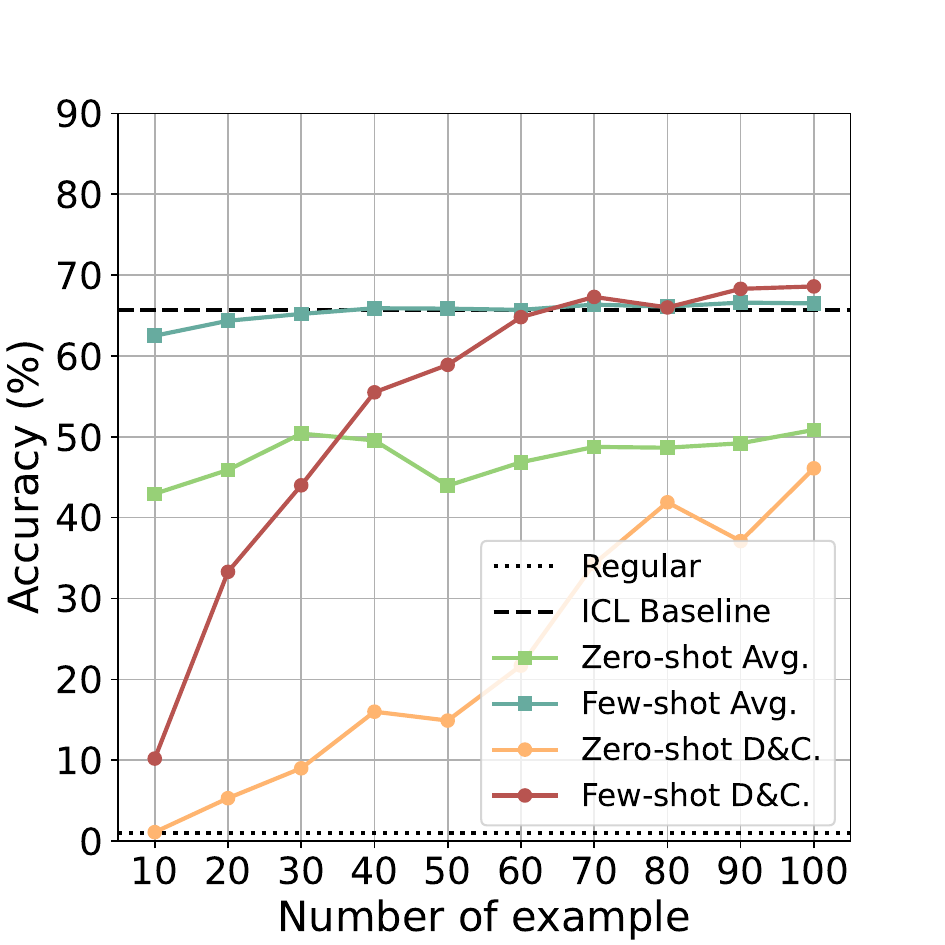}
    \captionsetup{font={scriptsize}}
    \caption{Antonym}
  \end{subfigure}
  \begin{subfigure}[b]{0.24\textwidth}
    \includegraphics[width=\linewidth]{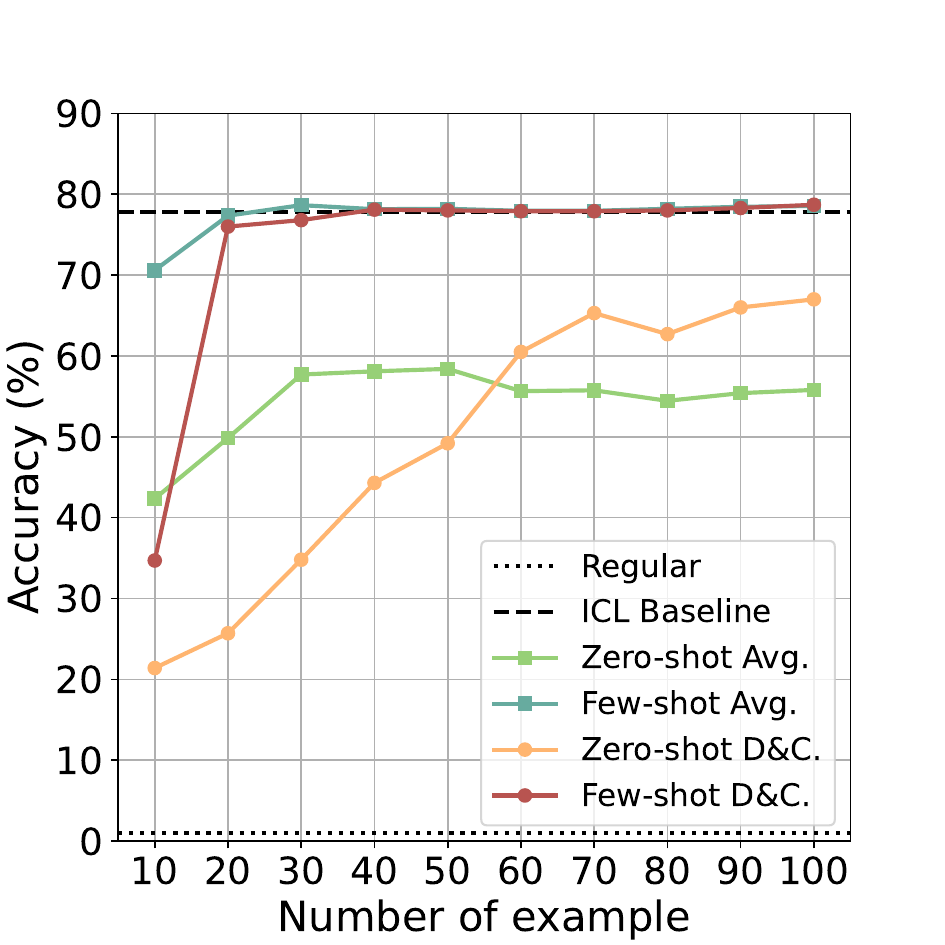}
    \captionsetup{font={scriptsize}}
    \caption{English-French}
    
  \end{subfigure}
  \begin{subfigure}[b]{0.24\textwidth}
    \includegraphics[width=\linewidth]{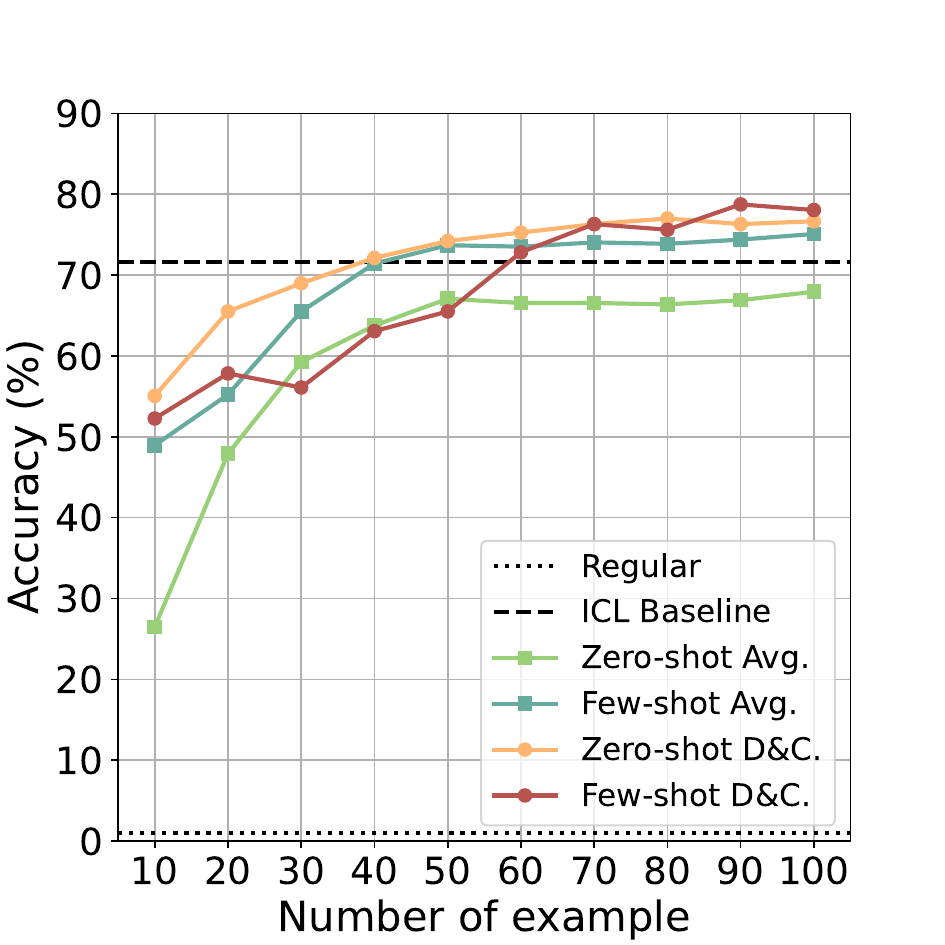}
    \captionsetup{font={scriptsize}}
    \caption{Product-Company}
    
  \end{subfigure}
  \caption{Performance of aggregation on Llama-2-13B across number of examples. \textit{Avg.} denotes the average aggregation baseline and \textit{D\&C.} denotes the divide-and-conquer aggregation. The \textbf{X} axis represents the number of examples, and the \textbf{Y} axis represents the accuracy.}
  \label{fig:aggregation_appendix_larger}
\end{figure*}

In this section, we provide the optimization and aggregation results on the larger model. Here we choose Llama-2-13B as its memory requirements suit our hardware conditions. We present the result of the optimization method on three representative datasets shown in Table~\ref{tab:main result appendix llama13B}, and the result of the aggregation method on four representative datasets shown in Figure~\ref{fig:aggregation_appendix_larger}. The result shows that our inner and momentum optimization and D\&C aggregation method could also benefit the state vector on the larger model setting. 

\section{Qualitative Study}

\begin{figure}[t]
    \centering
    \begin{subfigure}[b]{0.4\linewidth}
        \includegraphics[width=\linewidth]{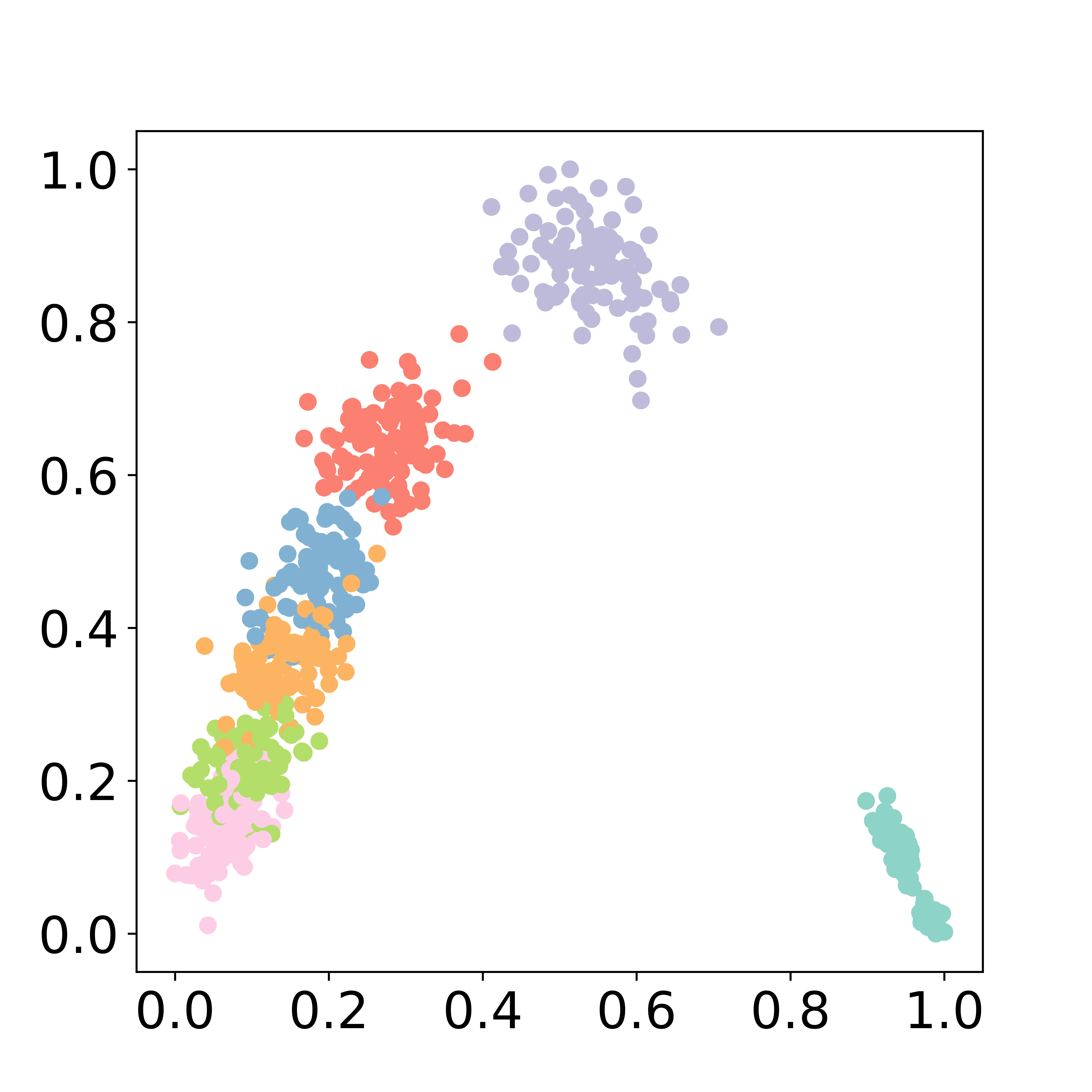}
        \caption{Antonym}
    \end{subfigure} 
    \begin{subfigure}[b]{0.4\linewidth}
        \includegraphics[width=\linewidth]{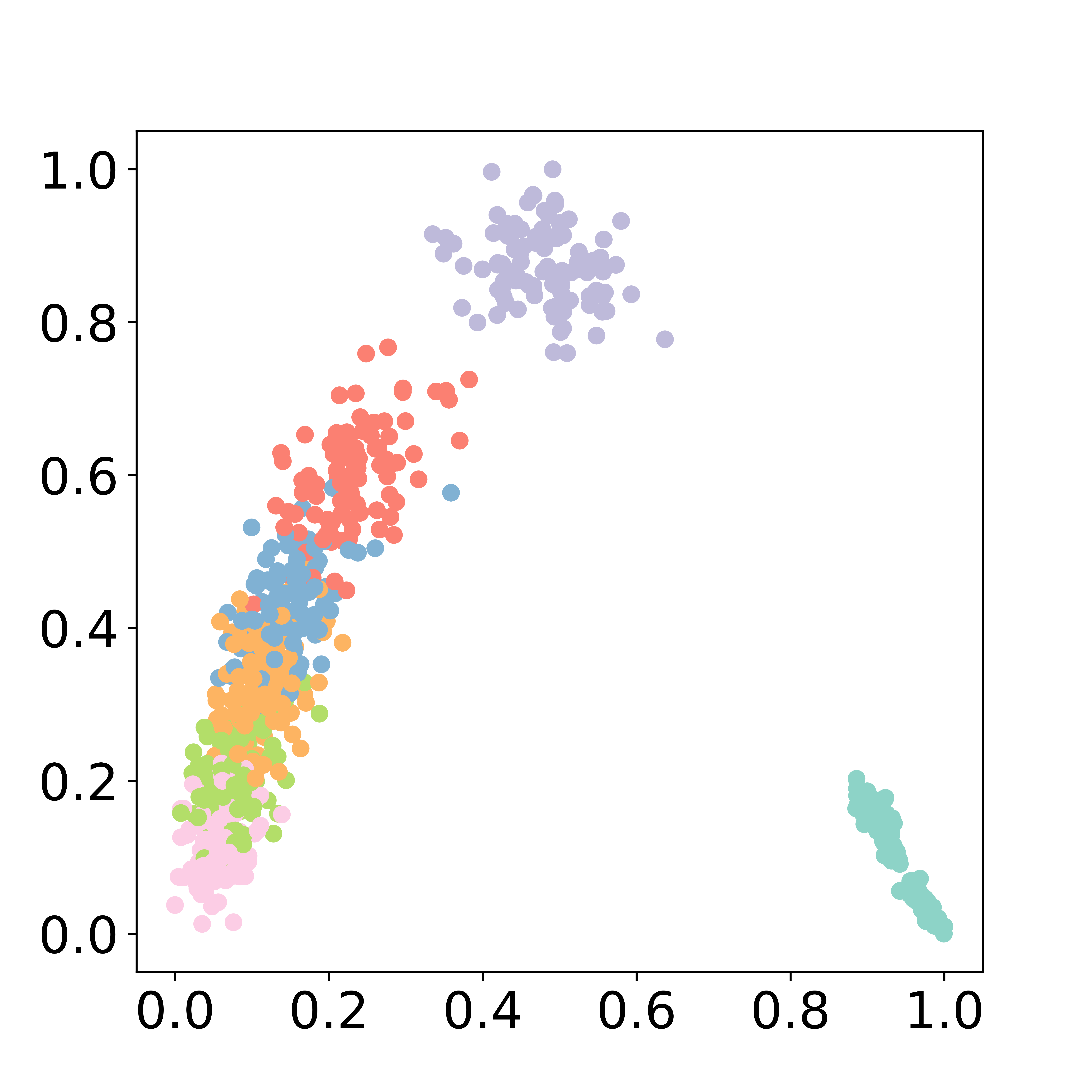}
        \caption{English-French}
    \end{subfigure} 
    
    \caption{The 2D PCA visualization of the state vector in the Antonym task and English-French task of \textbf{GPT-J}, where each color represents the state vector corresponding to examples occupying specific positions in the demonstration and the outlier is of the first order.}
    \label{fig:qualitative_appendix}
\end{figure}

In Figure~\ref{fig:qualitative_appendix}, we present a Principal Component Analysis (PCA) visualization of the original state vector in GPT-J, applied to both the Antonym task and the English-French translation task. Note that the cluster distributions observed in GPT-J closely mirror those of Llama-2. This similarity indicates a consistent and progressive enhancement in the model capacity, as originally identified in Llama-2 in \S\ref{sec: Qualitative Study}, which is also shown on GPT-J. Such findings demonstrate the broad applicability and generalizability of our momentum optimization approach across different models.

\begin{figure}[t]
\centering
\begin{subfigure}[b]{0.4\linewidth}
    \includegraphics[width=\linewidth]{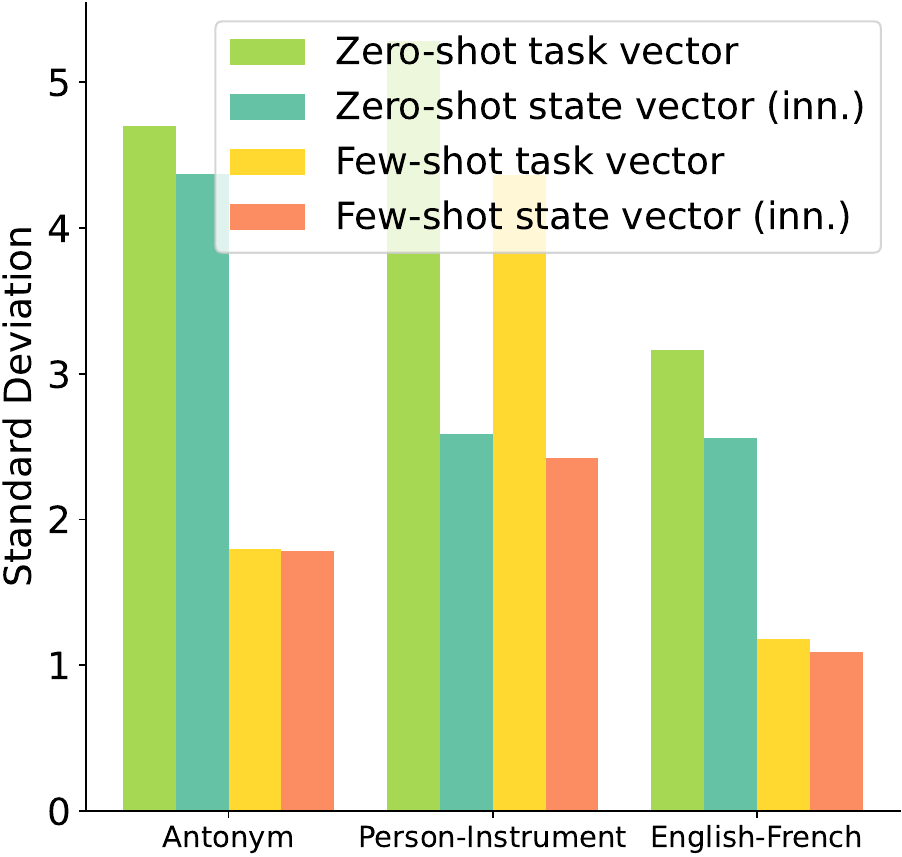}
    \caption{Demonstration Robustness}
  \end{subfigure}
  \begin{subfigure}[b]{0.4\linewidth}
    \includegraphics[width=\linewidth]{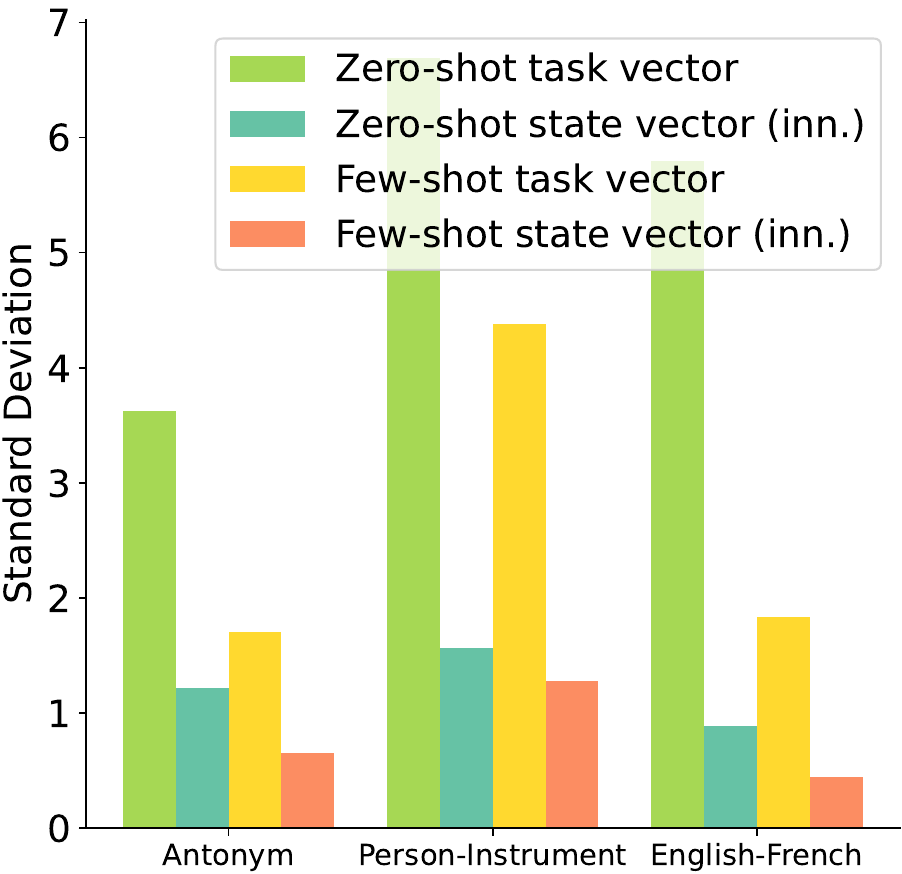}
    \caption{Dummy Robustness}
  \end{subfigure}
  \caption{Standard deviation of performance on Llama-2 across three datasets.}
  \label{fig:robust}
\end{figure}

\section{Robustness Analysis}

In this appendix, we examine the robustness of the state vector with inner optimization. Specifically, we evaluate the task vector and the inner optimized state vector on the Llama-2 dataset, focusing on three tasks. We measure and report the performance standard deviation using 100 diverse demonstrations or dummy queries. As illustrated in Figure~\ref{fig:robust}, our analysis yields three key observations:

\begin{itemize}
\item The task vector and state vector exhibit greater sensitivity to dummy queries than to demonstrations. This finding suggests that dummy queries have a greater impact on performance compared to demonstrations, underscoring the importance of reducing the noise from dummy queries to enhance state vector performance.
\item In the few-shot setting, both the task vector and the state vector (inn.) indicate significantly greater robustness compared to their performance in the zero-shot setting. There is a noticeable reduction in the standard deviation across diverse demonstrations or dummy queries when applying demonstrations during ICL inference. This improvement may be attributed to the richer ICL function information provided by demonstrations, which in turn bolsters performance stability.
\item Compared to the task vector, our inner optimized state vector shows markedly enhanced robustness to the variations in demonstrations and dummy queries, in both zero-shot and few-shot settings. This highlights the effectiveness of our proposed inner optimization in improving the robustness of the state vector.
\end{itemize}

\section{Limitation}
\label{sec:limitation}

The definition of state vectors is contingent upon specific assumptions and lacks a rigorous theoretical foundation, which may impact its generalizability and reliability across different NLP tasks. Additionally, the experiments were conducted on a limited scale with moderate-sized models and datasets. These constraints may affect the applicability of the results to larger models or more complex datasets. Further research will explore these aspects to establish a more robust validation of the proposed methods.

\end{document}